%% file: main.tex
\documentclass{ieeetj}
\usepackage{amsmath,amssymb,amsfonts}
\input{./latexGoodPractices/preamble}

\usepackage{algorithmic}
\usepackage{graphicx,color}
\usepackage{textcomp}
\usepackage{xcolor}
\usepackage{algorithm}
\def\BibTeX{{\rm B\kern-.05em{\sc i\kern-.025em b}\kern-.08em
    T\kern-.1667em\lower.7ex\hbox{E}\kern-.125emX}}
\AtBeginDocument{\definecolor{tmlcncolor}{cmyk}{0.93,0.59,0.15,0.02}\definecolor{NavyBlue}{RGB}{0,86,125}}

\usepackage{mathrsfs}

\acrodef{SLAM}{Simultaneous Localization And Mapping}
\acrodef{VSLAM}{Visual Simultaneous Localization And Mapping}
\acrodef{VO}{Visual Odometry}
\acrodef{AE}{Automatic-Exposure}
\acrodef{CRF}{Camera Response Function}
\acrodef{DN}{Digital Number}
\acrodef{SNR}{Signal-to-Noise Ratio}
\acrodef{HDR}{High Dynamic Range}
\acrodef{IMU}{Inertial Measurement Unit}
\acrodef{FPS}{frames per second}
\acrodef{API}{Application Programming Interface}
\acrodef{GP}{Gaussian Process}
\acrodef{RMSE}{Root-Mean-Square Error}
\acrodef{ICP}{Iterative Closest Point}
\acrodef{SIFT}{Scale-Invariant Feature Transform}
\acrodef{RPE}{Relative Pose Error}
\acrodef{RTE}{Relative Translation Error}
\acrodef{RRE}{Relative Rotation Error}
\acrodef{APE}{Absolute Pose Error}
\acrodef{GNSS}{Global Navigation Satellite System}
\acrodef{PoE}{Power over Ethernet}
\acrodef{RTK}{Real-Time Kinematic}
\acrodef{WILN}{Weather-Invariant Lidar-based Navigation}
\acrodef{ROS}{Robot Operating System}
\acrodef{RANSAC}{RANdom SAmple Consensus}

\def\numberTrajectories{59}
\def\kilometerTravelled{13.4}
\def\totalHours{\num{6.8}}
\def\numberImages{\num{538560}}
\def\numberExposureBracketing{six}

\def\selectionMethod{\textsc{HigherNoSat}}
\def\datasetName{BorealHDR}

\def\BracketList{\Delta T_\text{bracket}}
\def\BracketListExtension{\Delta T_\text{bracket\_ext}}
\newcolumntype{P}[1]{>{\centering\arraybackslash}m{#1}}
\newcolumntype{L}[1]{>{\raggedright\arraybackslash}m{#1}}

\def\OJlogo{\vspace{-4pt}$<$Society logo(s) and publication title will appear here.$>$}
\def\seclogo{\vspace{10pt}$<$Society logo(s) and publication title will appear here.$>$}

\def\authorrefmark#1{\ensuremath{^{\textbf{#1}}}}

\begin{document}
\receiveddate{XX Month, XXXX}
\reviseddate{XX Month, XXXX}
\accepteddate{XX Month, XXXX}
\publisheddate{XX Month, XXXX}
\currentdate{XX Month, XXXX}
\doiinfo{XXXX.2022.1234567}

\markboth{}{Author {et al.}}

\title{Reproducible Evaluation of Camera Auto-Exposure Methods in the Field: Platform, Benchmark and Lessons Learned}

\author{Olivier Gamache\authorrefmark{1}, Jean-Michel Fortin\authorrefmark{1}, Mat\v ej Boxan\authorrefmark{1} \\ François Pomerleau\authorrefmark{1}, and Philippe Giguère\authorrefmark{1}}
\affil{Université Laval, Québec, Canada}
\corresp{Corresponding author: Olivier Gamache (email: olivier.gamache@norlab.ulaval.ca).}
\authornote{This research was supported by Fonds de Recherche du Québec Nature et technologies (FRQNT) grant 334388, and Collaborative Research and Development Grant – Project (CRDPJ) entitled "Automation of Basic Forestry Operations" CRDPJ 538321 – 18, with FPInnovations, Produits forestiers Résolu inc.}

\begin{abstract}
Reproducibility is a cornerstone of scientific progress, as it enables fair comparisons between algorithms through the development of detailed solutions and datasets. 
However, standard datasets often present limitations, particularly due to the fixed nature of input data sensors, which makes it difficult to compare methods that actively adjust sensor parameters to suit environmental conditions. 
This is the case with \ac{AE} methods, which rely on environmental factors to influence the image acquisition process. 
As a result, \ac{AE} methods have traditionally been benchmarked in an online manner, rendering experiments non-reproducible. 
Building on our previous work~\citep{gamache2024exposing}, we propose a methodology that utilizes an emulator capable of generating images at any exposure time. 
This approach leverages \datasetName, a unique multi-exposure stereo dataset, along with its new extension, in which data was acquired along a repeated trajectory at different times of the day to assess the impact of changing illumination. 
In total, \datasetName~covers \SI[detect-weight=true,mode=text]{\kilometerTravelled}{\kilo\meter} over \si[detect-weight=true,mode=text]{\numberTrajectories} trajectories in challenging lighting conditions. 
The dataset also includes lidar-inertial-odometry-based maps with pose estimation for each image frame, as well as \ac{GNSS} data for comparison. 
We demonstrate that by using images acquired at various exposure times, we can emulate realistic images with a \ac{RMSE} below \SI[detect-weight=true,mode=text]{1.78}{\percent} compared to ground truth images. 
Using this offline approach, we benchmarked eight \ac{AE} methods concluding that the classical \ac{AE} method remains the field's best performer. 
To further support reproducibility, we provide in-depth details on the development of our backpack acquisition platform, including hardware, electrical components, and performance specifications. 
Additionally, we share valuable lessons learned from deploying the backpack over more than \SI{25}{\kilo\meter} across various environments.
Our code and dataset are available online at this link: \url{https://github.com/norlab-ulaval/TFR24_BorealHDR}
\end{abstract}

\begin{IEEEkeywords}
Computer vision, Image processing, Mobile robots, Robot sensing systems, Visual odometry
\end{IEEEkeywords}


\maketitle

\begin{figure}[tbp] 
	\centering
	\includegraphics[width=0.48\textwidth]{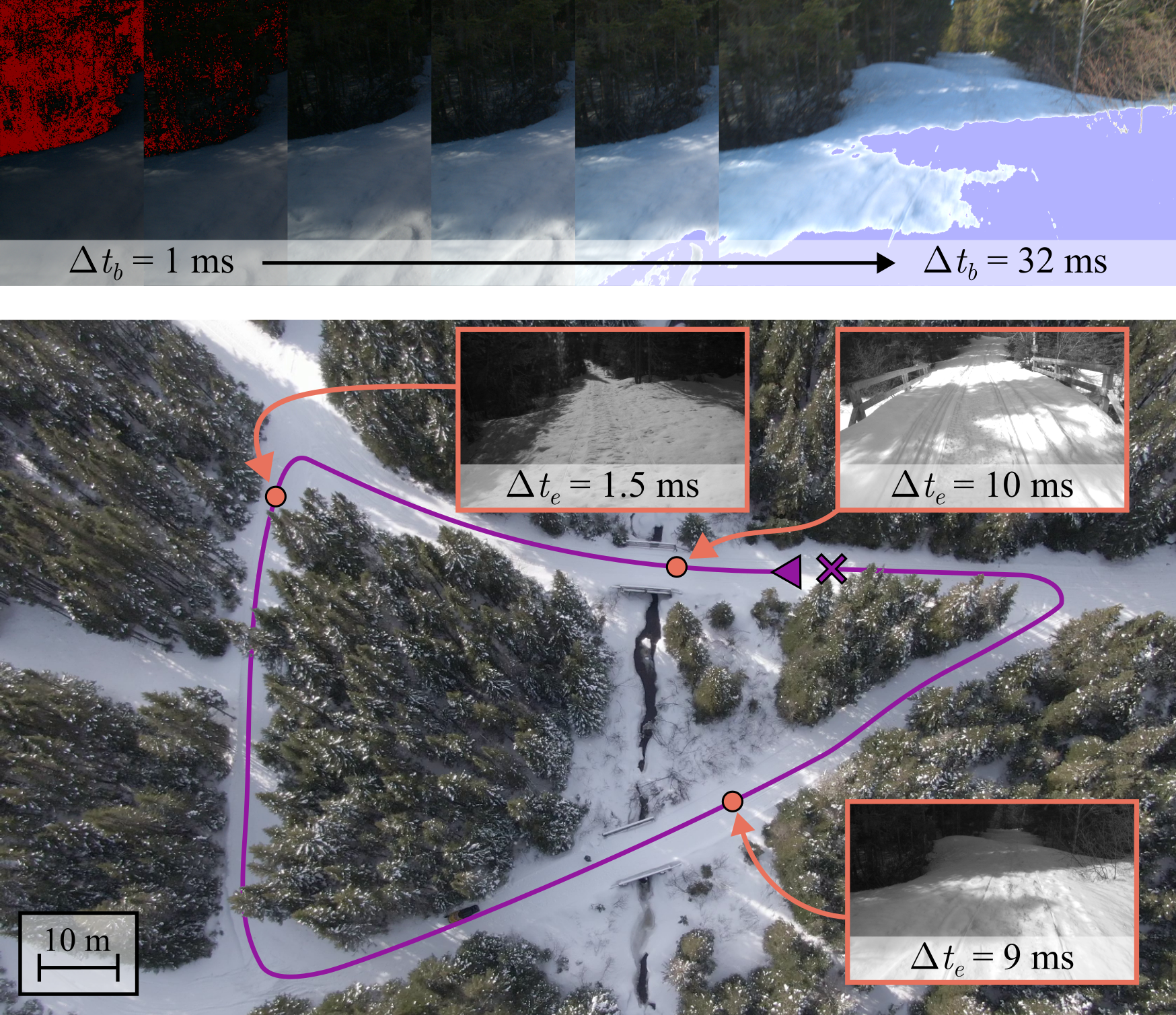}
	\caption{
            \textit{Upper image}: Example of the six acquired brackets, with one used to generate \bm{$\Delta t_e =$} \SI[detect-weight=true,mode=text]{9}{\milli\second} from the lower image.
            The bracket exposure times \bm{$\Delta t_b$} are \bm{$\{1, 2, 4, 8, 16, 32\}$} \si[detect-weight=true,mode=text]{\milli\second}, increasing the dynamic range of our capture by 30 dB or five stops. 
            Blue and red colorized pixels are over-exposed and under-exposed, respectively.
            \textit{Lower image}: Overhead view of a trajectory from our \datasetName~dataset, taken in Montmorency Forest, Québec, Canada.
            The traveled trajectory is shown in purple. 
            Possible emulated exposure times \bm{$\Delta t_{e}$} along the trajectory are depicted in orange.
            }
	\label{fig:intro} 
\end{figure}

\section{INTRODUCTION}
Experiment reproducibility represents a significant challenge in field robotics \citep{lier2016towards}.
Detailed and replicable systems accelerate innovation by enabling fair comparisons with established methods.
In mobile robotics, reproducibility has been emphasized in several subdomains through the adoption of benchmarking datasets and frameworks, as seen in visual \ac{SLAM} systems.
For instance, datasets such as KITTI~\citep{KITTI2013} or Oxford RobotCar~\citep{maddern20171} have pushed forward the development of state-of-the-art visual localization algorithms by providing standardized environments for testing and comparison.

One major limitation of pre-recorded datasets is the lack of control over sensor measurements, which significantly impacts the performance of visual \ac{SLAM} algorithms \citep{liu2021simultaneous}.
Effective control of the camera's exposure parameters, in particular, is essential for detecting key features in the scene and maintaining accurate tracking.
These auto-exposure algorithms play a crucial role in visual \ac{SLAM}, especially in outdoor settings where \ac{HDR} scenes and rapidly changing illumination can severely compromise performance~\citep{Chahine2018}.
For instance, a car emerging from a tunnel may experience illumination variations exceeding \SI{120}{\decibel} within a second~\citep{Tomasi2021}, whereas a standard 12-bit camera provides a theoretical dynamic range of only \SI{72}{\decibel}.
Similarly, boreal forests during winter are highly-contrasted scenes, where sunlight reflected on snow contrasts sharply with shaded trees~\citep{Baril2022}.
This contrast inevitably leads to saturated pixels on both ends of the spectrum, as highlighted by blue and red colorized hues in~\autoref{fig:intro}, resulting in the loss of valuable information for visual \ac{SLAM} algorithms.

To address such challenges, researchers have developed \acf{AE} algorithms that dynamically adjust camera parameters, such as exposure time and gain, during operation to reduce the impact of the scene's dynamic range on visual \ac{SLAM}~\citep{Zhang2017,Kim2020,begin2022}.
However, the systematic evaluation of camera \ac{AE} algorithms remains largely unexplored and lacks standardized benchmarks.
Consistent evaluation is particularly challenging because exposure control operates in real-time~\citep{Shin2019,Shim2019}. 
This requires precisely replicating the test conditions for each run, including the lighting and camera poses.
In uncontrolled environments where lighting is constantly changing, like the one shown in~\autoref{fig:intro}, achieving reproducibility becomes virtually impossible.
While no standard benchmark has been set, authors of recent \ac{AE} methods often propose guidelines.
A common approach involves capturing images simultaneously with multiple cameras, one for each method~\citep{Shim2019}, but this becomes increasingly complex and expensive as the number of tested methods grows.
Another alternative involves replicating identical trajectories multiple times for each evaluated \ac{AE}~\citep{begin2022}, which becomes impractical for longer trajectories.
These limitations highlight the need for more practical and reproducible benchmarking solutions.


Building on our previous work~\citep{gamache2024exposing}, this paper introduces a framework for equitable comparisons of \ac{AE} algorithms applied to robot localization. 
Leveraging the bracketing technique~\citep{gupta2013fibonacci} and realistic image emulation, our method addresses many of the reproducibility challenges in \ac{AE} evaluation, by giving control over the input data.
In this extended study, we perform an in-depth and unbiased benchmarking of eight \ac{AE} methods using an extended version of the \datasetName~dataset~\citep{gamache2024exposing},
featuring a single trajectory repeated six times under varying lighting conditions throughout the day.
Additionally, we provide a complete description of our custom-made data collection system to encourage reproducibility.
This includes details about the system's electrical components, hardware configurations, and limitations, complemented by open-source software.
Finally, based on insights from kilometer-scale deployments of our platform, we share lessons learned that can lead to future developments in this domain and benefit the community.
In short, our contributions are the following:
\begin{enumerate}
    \item An offline emulation framework to generate photo-realistic images at any plausible exposure time based on our \datasetName~dataset and its new extension;
    \item In-depth analysis of the impact of \ac{AE} on visual \ac{SLAM} based on an unbiased offline benchmarking of eight methods;
    \item A detailed description of our versatile backpack, including hard-to-access details on the built-in components for better replicability; 
    \item Insights and lessons learned from multiple field deployments and from the development of a custom platform. 
\end{enumerate}

\section{RELATED WORK}
\label{sec:related_work}

We first review the developed methodologies for comparison of \ac{AE} algorithms and analyze their reproducibility potential.
Then, we go over available datasets for visual \ac{SLAM} algorithms and their limitations in the context of testing \ac{AE}, highlighting the advantages of our dataset.
After that, we detail the existing human-portable platforms and their use in the robotic community.
Finally, we discuss important state-of-the-art \ac{AE} algorithms that will be used in our thorough benchmarking.

\subsection{METHODOLOGIES FOR \ac{AE} ALGORITHMS COMPARISON}
\label{sec:related_work_methodology}

The nature of \ac{AE} algorithms makes their comparison very challenging, as they actively change camera parameters during execution.
Here, we present state-of-the-art \ac{AE} benchmarking methods that address this issue, and categorize them between static and moving cameras approaches. 

\textbf{Static Cameras --} To evaluate \ac{AE} methods, a common approach involves fixing a camera and capturing multiple images at varying exposure times. \citet{Shim2019} employed a surveillance camera to acquire images under 210 different exposure parameter combinations, repeated six times daily, to compare their proposed \ac{AE} method against existing techniques. By selecting the image closest to the desired outcome for each method, they demonstrated that their approach consistently chose images with the highest gradient levels.
Similarly, \citet{Zhang2017} introduced a dataset comprising 18 real-world static scenes, each acquired at multiple exposure levels. Their analysis showed that their gradient-based metric outperformed state-of-the-art methods, identifying more features over \SI{70}{\percent} of the images from their dataset. 
\citet{Kim2020} curated a dataset of \num{1000} images spanning various exposure settings, capturing both indoor and outdoor scenes.
Instead of using a monocular camera, \citet{Shin2019} utilized a stereo camera to construct a static scene dataset of 25 images. Their noise-aware \ac{AE} method provided images with reduced saturation compared to state-of-the-art techniques.
Datasets of multi-exposed static scenes are valuable for evaluating proxies of visual \ac{SLAM}, such as single-image feature detection. However, to benchmark \ac{AE} algorithms within complete visual \ac{SLAM} pipelines, datasets containing full trajectories, such as our proposed multi-exposure dataset, are essential.

\textbf{Moving Cameras --} Three trends can be observed using cameras in motion to compare \ac{AE} algorithms: \textit{simulation-based} methods, \textit{multi-camera} methods, and \textit{multi-trajectory} methods.

\emph{Simulation-based} methods, as the name suggest, rely on simulated data for synthetic evaluation of \ac{AE} algorithms.
For instance, such comparison was made by ~\citet{Zhang2017} through the Multi-FoV synthetic dataset ~\citep{zhang2016benefit}, to simulate multiple versions of an image, but with different exposition levels.
Similarly, ~\citet{gomez2018learning} trained a neural network to produce images with higher gradient information using a synthetic dataset, where they simulated \num{12} different exposures.
Although these techniques reduce dataset development time, they face the \textit{Sim2Real} gap~\citep{hofer2021sim2real}, whereas our multi-exposure dataset provides all the advantages of offline evaluation while relying only on real-world images.

\emph{Multi-camera} methods are the most widespread for \ac{AE} algorithms comparison, leveraging a hardware solution for effective data collection.
In this setup, two or three cameras are typically installed on a mobile platform~\citep{Shim2019,Kim2020,Mehta2020} allowing exact comparison since all \ac{AE} methods are executed simultaneously, thus facing the same conditions.
\citet{Wang2022} used four stereo cameras, facilitating \ac{VO} comparison, as the stereo depth estimation provides scaling information.
An important drawback for these multi-camera methods is that the hardware complexity grows with the number of \ac{AE} methods tested, rapidly becoming impractical.
On the contrary, our emulation approach can be used to compare on an unlimited number of \ac{AE} methods, at a fixed acquisition cost. 

\emph{Multi-trajectory} methods consist of repeating multiple times the exact same trajectory using a different exposure control scheme at each iteration.
This technique was used by \citet{begin2022} to develop a gradient \ac{AE} technique, which was tested using two cameras installed on a motorized \SI{0.16}{\meter\squared} motion table.
Their setup allowed for exact repetitions of the same trajectories multiple times, with a ground truth precision near \SI{0.1}{\milli\meter}.
To evaluate \ac{AE} methods in the presence of motion blur, \citet{Han2023} acquired images from three cameras in motion on a motorized rail.
While providing precise ground truth, the rail approach severely restrains the total area covered by the trajectories.
It also limits the number of observable environments.
A variation of the \textit{multi-trajectory} method was developed by~\citet{Kim2020}.
They drove using stop-and-go maneuvers on a single trajectory, where they stopped six times in total.
At each stop, they collected images at multiple exposure times.
This technique restricts the number of images that can be acquired, since any changes in the environment would corrupt the benchmark.

Our proposed comparison approach does not make any static assumption, thanks to the bracketing acquisition technique and our emulation pipeline.
We are able to compare \ac{AE} methods using standard visual \ac{SLAM} pipelines on dynamic trajectories.
Inspired by our previous work~\citep{gamache2024exposing}, \citet{zhang2024efficient} collected a bracketing dataset comprising four trajectories, enabling the emulation of various image conditions. 
Additionally, they replicated two state-of-the-art exposure control methods to compare with their own \ac{AE} algorithm.
In contrast, our evaluation is completely impartial, as we only assess exposure control methods that were developed independently of our research.

\subsection{DATASETS FOR VISUAL \ac{SLAM}} 
\label{sec:related_work_dataset}

Several datasets were collected aiming at improving visual \ac{SLAM} against challenging illumination conditions.
The KITTI-360~\citep{liao2022kitti} and the Oxford~\citep{Oxford2020} datasets both acquired multiple cameras images, lidars, \ac{IMU}, and \acf{GNSS} data, with Oxford offering different weathers, seasons, and illuminations.
The North Campus dataset~\citep{NorthCampus} also acquired urban stereo images, changing between indoors and outdoors, on a 15-month range, using a segway robotic platform.
The UMA-VI dataset~\citep{UMA-VI} had for main purpose to acquire \ac{HDR} images with a large number of low-textured scenes.
They used a handheld custom rig equipped with cameras and \ac{IMU}, but they only provided ground truth through loop closure error.
Closer to our dataset environment, TartanDrive~\citep{TartanDrive2022} and FinnForest~\citep{FinnForest2020} both acquired off-road data.
The TartanDrive dataset contains seven proprioceptive and exteroceptive modalities used for reinforcement learning purposes, including stereo images.
Although simpler, the FinnForest dataset is also composed of stereo images in summer and winter, showing the same forest landscapes under multiple conditions.
In the context of challenging environments for \ac{AE} methods, the Visual-Inertial Canoe Dataset~\citep{miller2018visual} provides \ac{GNSS} measurements alongside stereo camera images from a canoe navigating a river.
The platform being on water introduces complex conditions such as multiple reflections and lens flare, which pose significant challenges for visual \ac{SLAM} algorithms.

From the papers described in \autoref{sec:related_work_methodology}, only \citet{begin2022} and \citet{Shin2019} published their dataset.
While the existing datasets allowed great improvements of visual \ac{SLAM} algorithms, ours is complementary by providing a large dynamic range of scenes through exposure times cycling. 
Combined with our emulation framework, we unlock the possibility to select the exposure time \emph{during playback}, expanding the realm of camera parameters algorithms evaluation.
In this paper, as outlined in our contributions, we extend our multi-exposure \datasetName~dataset to a total of \num{\numberTrajectories}~trajectories for a more realistic evaluation of the \ac{AE} methods in the field.

\subsection{PORTABLE ACQUISITION PLATFORMS}
\label{sec:related_work_platforms}



Standard motorized acquisition platforms designed for outdoor navigation are often unsuitable for narrow or constrained environments~\citep{knights2023wild}.
Consequently, for tasks that rely solely on sensor-based data collection, a wearable system offers superior versatility and agility to navigate and adapt to such settings. 
This is the case of \citet{duan2023low} and \citet{zhang2024efficient}, where they both designed handheld devices that incorporate multiple sensors, such as lidar, \ac{IMU} and/or cameras.
The Wild-Places dataset~\citep{knights2023wild} was also collected using a handheld sensor payload device, where they recorded an impressive \SI{33}{\kilo\meter} of camera, lidar, and \ac{IMU} data.
Most of the time, handheld devices have low computing power and battery life, because of the limitation in weight of the platform~\citep{shi2022polyu}.
Another format of human-portable system is the backpack-type platforms, which are more common because of their ergonomic characteristics.
Several research groups~\citep{chen2021low, bao2022systematic, zhou2023backpack} developed their own backpack version for lidar mapping experiments.
Similar to our backpack, \citet{shi2022polyu} developed a system with multiple modalities, such as two lidars, a \SI{360}{\degree} camera, an \ac{IMU}, and a \ac{GNSS} module.
They give several insights on their sensors and the data format of every sensors in the dataset.
Their system operates on Li-ion batteries, providing a \SI{5}{\hour} runtime, which supports the sensors and an onboard Intel NUC i7 computer.
However, the level of detail provided for the described solutions is insufficient to enable the reproduction of any of the designs.
In~\autoref{sec:experimental_setup_backpack}, we address this by providing the electrical configuration, sensor specifications, 3D model, and implemented software, ensuring that our backpack system is fully replicable by the research community.

\subsection{AUTOMATIC EXPOSURE ALGORITHMS}
\label{sec:related_work_ae}

We look at existing \ac{AE} methods from the state-of-the-art and evaluate the most relevant to our study.
One of the most important aspect of vision-based localization is the necessity of sufficient contrast in images to maintain accuracy and robustness.
The contrast level, in turns, depends heavily on the camera's \ac{AE} algorithm and its accompanying target metric.
For instance,~\citet{Shim2019} used image gradient magnitude as such metric. 
Their exposure control scheme generates seven synthetic versions of the latest acquired image, simulating different exposure levels, to identify the next exposure value maximizing their metric.
The \ac{AE} algorithm proposed by~\citet{Zhang2017} instead sorts the gradient level of each pixel, and applies a weight factor based on their percentile value.
By combining their quality metric and the \ac{CRF}, they predict the best next exposure value.
Largely based on~\citep{Zhang2017}, \citet{Wang2022} also uses the \ac{CRF} to estimate the photometric most sensitive region, meaning the pixel intensities that exhibit the greatest response to changes in illumination.
Additionally, they simulate images on either side of the calculated optimal exposure time and leverage these to determine the direction in which the exposure time should adjust to optimize their gradient-based metric.
Using this search-based strategy, their method demonstrates rapid convergence following abrupt illumination variations.
\citet{Kim2020} developed an image quality metric based on the gradient level and Shannon entropy~\citep{shannon1948mathematical}, used to detect saturation.
Recently, \citet{zhang2024efficient} introduced a solution based on deep reinforcement learning.
Influenced by the bracketing data acquisition from our previous work \citep{gamache2024exposing}, they trained their network using emulated images as self-annotated data.
They combined the number of detected features in the image $\mathscr{R}_{\text{detect}}$ and the number of matches with consecutive images $\mathscr{R}_{\text{match}}$ to build a reward function aiming at improving visual \ac{SLAM} performances:
\begin{equation}
    \mathscr{R}_{\text{feat}} = \mathcal{W}_{\text{detect}} \cdot \mathscr{R}_{\text{detect}} + \mathcal{W}_{\text{match}} \cdot \mathscr{R}_{\text{match}},
\label{eq:reward_fct}
\end{equation}
where $\mathcal{W}_{\text{detect}}$ and $\mathcal{W}_{\text{match}}$ are scaling coefficients.
Their exposure control methods achieved good accuracy, mostly in situations where the illumination changed radically.
To demonstrate the benchmarking capabilities of our approach, we provide an implementation of the above methods, which are often used for comparison~\citep{begin2022,Wang2022, zhang2024efficient}, since they cover main aspects of image quality for localization algorithms. 
In this paper, we focus on \ac{AE} methods primarily driven by exposure time control, in contrast to other approaches, such as \citet{Shin2019} which takes into account the \ac{SNR} and rely mostly on gain control.

\section{METHODOLOGY}
\label{sec:theory}

To reinforce our primary contribution, we present an emulation framework capable of generating reproducible images at any realistic exposure time in an offline setting. 
This section outlines the approaches adopted for the development of our system, focusing on the emulation technique and the benchmarking metrics used to evaluate \ac{AE} methods.

\subsection{EMULATION TECHNIQUE}
\label{sec:theory_emulation}

Our emulator leverages the bracketing photography technique, which involves varying the exposure time for each captured image.
Therefore, by starting from a small exposure time to a long one, we cover the whole dynamic range of a scene.
The resulting sequence of images, $\{I_i, \Delta t_i\}_{i=1}^N$ serves as input to the emulator, where $I_i$ is the image, $\Delta t_i$ is the corresponding exposure time, and $N$ is the number of acquired exposure times in one full bracketing cycle.

The main idea behind our emulation method is based on the image acquisition process, which maps the scene radiance $E$ to the image pixel values $I(x)$, using the vignetting function $V(x)$, the exposure time $\Delta t$, and the \ac{CRF} $f(\cdot)$.
This process can be expressed, based on~\citep{Bergmann2018}, using the following equation:
\begin{equation}
    I(x) = f\left(\Delta t V(x) E \right).
    \label{eq:photometric}
\end{equation}
From~\autoref{eq:photometric}, the relationship between two images $I_\text{source}$ and $I_\text{target}$ and their respective exposure times $\Delta t_\text{source}$ and $\Delta t_\text{target}$ can be defined as
\begin{equation}
    I_\text{target} = f\left(\frac{\Delta t_\text{target}}{\Delta t_\text{source}}\cdot f^{-1}\left(I_\text{source}\right)\right),
    \label{eq:emulation}
\end{equation}
where $f^{-1}(\cdot)$ is the inverse \ac{CRF}~\citep{Grossberg2003}.
The \ac{CRF} $f(\cdot)$ and its inverse $f^{-1}(\cdot)$ can be estimated from multiple images of a static scene at several exposure times, allowing to capture the whole dynamic range at fixed radiance, following~\citep{engel2016monodataset}.
With $f^{-1}(\cdot)$ and~\autoref{eq:emulation}, it is thus possible to emulate any targeted exposure time $\Delta t_\text{target}$ by using a known image with an exposure time $\Delta t_\text{source}$.
Note that, for the sake of simplicity, an image taken from one exposure time in the bracketing cycle will be called \textit{bracket}.

Based on \autoref{eq:emulation}, we propose an emulation algorithm, detailed in this section and in \textcolor{red}{Algorithm} \autoref{alg:pseudo_code}. 
Considering that the data used for the emulation is acquired while moving, the brackets are not taken at the same time nor position. 
To avoid introducing artifacts into the resulting image, we opted against interpolating between brackets.
Instead, we select the most appropriate $I_\text{source}$ from the available brackets $\{I_i, \Delta t_i\}_{i=1}^N$ to emulate $I_\text{target}$.
The selection process involves two key considerations. First, the distance between $\Delta t_{\text{source}}$ and $\Delta t_{\text{target}}$ is evaluated.
To maximize the \ac{SNR} in the emulated image, we prioritize using a $\Delta t_{\text{source}}$ such as $\Delta t_\text{target}/\Delta t_\text{source} < 1$, ensuring pixel values are scaled down.
Second, high exposure times can increase image saturation.
Given the limited dynamic range of standard cameras, excessively bright or dark pixels become saturated and lose meaningful information.
To mitigate this, we select $I_{\text{source}}$ such that its saturation level remains below a predefined threshold $\alpha$.




Based on these considerations, we developed a selection method named \selectionMethod.
The first step is to find the two closest brackets $\Delta t_\text{bl}$ and $\Delta t_\text{bh}$, that bound $\Delta t_\text{target}$, such that
\begin{equation}
    \Delta t_\text{bl} < \Delta t_\text{target} < \Delta t_\text{bh}.
    \label{eq:bound}
\end{equation}
Then, we compute the amount of image saturation in the bracket associated with $\Delta t_\text{bh}$, where the saturation corresponds to the number of pixels with values of \num{0} and \num{4095}, for our \num{12}-bit channel images.
If the saturation level of the higher bracket is below $\alpha$, we select it as the best candidate, otherwise, we pick the lower one.
Finally, if $\Delta t_\text{target}$ is outside the range of available $\Delta t_\text{source}$, we select the closest bracket.

\begin{algorithm}
\caption{Exposure Emulation Using Bracketing}
\begin{algorithmic}[htbp]
\STATE \textbf{Input:}\\
\hspace{0.3cm}Brackets: $\{I_i, \Delta t_i\}_{i=1}^N$ \\
\hspace{0.3cm}Target exposure time: $\Delta t_{\text{target}}$
\STATE \textbf{Output:}\\ 
\hspace{0.3cm}Emulated image: $I_{\text{target}}$
\STATE \textbf{Main:}\\ 
\STATE Identify the closest brackets $\Delta t_{\text{bl}}$ and $\Delta t_{\text{bh}}$, such that:
\[
\Delta t_{\text{bl}} < \Delta t_{\text{target}} < \Delta t_{\text{bh}}
\]

\IF{Saturation level of $\Delta t_{\text{bh}} < \alpha$}
    \STATE $I_{\text{source}} = I_{\text{bh}}$, $\Delta t_{\text{source}} = \Delta t_{\text{bh}}$
\ELSE
    \STATE $I_{\text{source}} = I_{\text{bl}}$, $\Delta t_{\text{source}} = \Delta t_{\text{bl}}$
\ENDIF

\IF{$\Delta t_{\text{target}} \notin \left[\Delta t_0, \Delta t_N\right]$}
    \STATE Select the closest bracket.
\ENDIF

$I_{\text{target}} = f\left( \frac{\Delta t_{\text{target}}}{\Delta t_{\text{source}}} \cdot f^{-1}(I_{\text{source}}) \right)$

\STATE \textbf{Return:} $I_{\text{target}}$
\end{algorithmic}
\label{alg:pseudo_code}
\end{algorithm}


\subsection{BENCHMARKING EXPERIMENTS}
\label{sec:theory_experiments}


To conduct a thorough evaluation of existing \ac{AE} methods, we design experiments that emphasize critical components of visual \ac{SLAM} algorithms. 
The analysis focuses on feature quantity and distribution, as well as performance when tested on a state-of-the-art visual \ac{SLAM} algorithm.
This section describes the evaluated metrics that are used in section \autoref{sec:results}.

\subsubsection{Feature Tracking and Uniformity}
\label{sec:theory_experiments_features}

A widely used approach to assess image quality for visual \ac{SLAM} algorithms involves detecting features and calculating associated metrics, such as the number of features detected~\citep{Zhang2017,Shin2019}.
In addition to the number matches, we also analyze the spatial distribution of features in images for each of the benchmarked \ac{AE} methods. 
A uniform feature distribution is considered a good proxy for visual \ac{SLAM} algorithm performance.
To measure this, each image is divided into a $n \times n$ grid, and a cell is marked as "filled" if at least one feature is detected within it. 
Therefore, a higher count of filled cells indicates better feature coverage across the image.

\subsubsection{Relative Trajectory Error}
\label{sec:theory_experiments_vslam}

A key advantage of our emulation-based approach is its ability to reconstruct traveled trajectories at various exposure times, enabling the generation of images during post-processing. 
Using our proposed framework, we emulate images along a trajectory based on the specific exposure times desired by each \ac{AE} method, and subsequently feed these images into a state-of-the-art visual \ac{SLAM} algorithm.
To assess the accuracy of the resulting trajectories, we compare them to a reference trajectory, and calculate the \ac{RTE} and \ac{RRE} based on~\citep{sturm2012benchmark}.
First, we compute the \ac{RTE} for a single trajectory $t$ using
\begin{equation}
    \text{RTE}(t, w) = \sqrt{\frac{1}{S}\sum^S \left(\frac{E_{T,S}}{w} \cdot 100 \right)^2},
    \label{eq:rte}
\end{equation}
where $w$ is the evaluation window size, $S$ is a segment of size $w$ from the trajectory, and $E_{T,S}$ is the translation error between the reference and the evaluated trajectory on the segment $S$.
To estimate the average \ac{RTE} on all the trajectories $T$, we compute
\begin{equation}
    \text{RTE}\left[\si{\percent}\right] = \frac{1}{T}\frac{1}{\text{len}(\bm{W})} \sum_{t=1}^T \sum_{i=1}^{\text{len}(\bm{W})}\text{RTE}(t,\bm{W}(i)),
    \label{eq:RTE}
\end{equation}
where $\bm{W}$ is a vector containing all the evaluated window values $w$.
Similar to \autoref{eq:rte} and \autoref{eq:RTE}, we repeat the same steps for the \ac{RRE}, which is given by 
\begin{equation}
    \text{RRE}(t, w) = \sqrt{\frac{1}{S}\sum^S \left(\frac{E_{R,S}}{w} \right)^2},
    \label{eq:rre}
\end{equation}
and
\begin{equation}
    \text{RRE}\left[\si{\degree/\meter}\right] = \frac{1}{T}\frac{1}{\text{len}(\bm{W})} \sum_{t=1}^T \sum_{i=1}^{\text{len}(\bm{W})}\text{RRE}(t,\bm{W}(i)).
\end{equation}
In \autoref{eq:rre}, $E_{R,S}$ is the rotational error on the segment $S$.
Note that the relative errors are divided by $w$ to reduce the impact of the increasing error when comparing on longer trajectory segments.

\subsubsection{Robustness Evaluation}
\label{sec:theory_experiments_robustness}

While the previous metrics focus on the accuracy, the following evaluate the robustness of the different approaches.
Similar to the failure metric describes in~\citep{naveed2022deep}, we measure the duration of each trajectory before failure when using the visual \ac{SLAM} algorithm.
Evaluating this metric across multiple trajectories provides insight into which \ac{AE} methods select exposure times that enable more consistent navigation.
Additionally, we compute the number of successful trajectories achieved by the visual \ac{SLAM} algorithm, where a successful trajectory is defined as one in which the algorithm maintains localization without becoming lost for the whole trajectory.
Finally, we leverage the extended capabilities of our \datasetName~dataset to examine the effects of changing illumination over the course of an entire day. 
Using emulated images captured at the same location along the repeated trajectory, we illustrate how different \ac{AE} methods adapt to varying lighting conditions.

\subsubsection{Statistical Analysis of Automatic-Exposure Methods}
\label{sec:theory_experiments_statistical}

The final metric aims to support our second contribution, where we assess whether state-of-the-art \ac{AE} methods outperform a standard camera’s automatic exposure when deployed in the field. 
For each of the metrics described above, we analyze the distribution of values across all trajectories and perform a statistical hypothesis test. 
To avoid assumptions about the distribution shape, we apply a non-parametric test with a $1 - \beta$ confidence interval, using the Mann-Whitney \textit{U} test \citep{mann1947test}.
This test is conducted between the distributions of the benchmarked methods and the classical automatic exposure approach.
The null hypothesis posits that the distributions are similar, while the alternative hypothesis suggests they differ.
Based on the results of this hypothesis test, we retain the \ac{AE} methods that differ from the reference method, and conduct the Mann-Whitney \textit{U} test a second time to determine if their performance is significantly better or worse than the classical automatic exposure method.
In this case, the null hypothesis assumes the distributions are similar, and the alternative hypothesis suggests the evaluated \ac{AE} methods outperform the classical automatic exposure approach.
Considering multiple hypothesis tests are performed using the same reference distribution, the probability of rare events increases, potentially leading to a false rejection of the null hypothesis.
To mitigate this risk, we apply a Bonferroni correction \citep{bonferroni1936teoria} to adjust the threshold value
\begin{equation}
    \beta \leftarrow \frac{\beta}{n_{test}},
\label{eq:bonferroni}
\end{equation}
where $n_{test}$ if the number of \ac{AE} methods on which we perform the hypothesis test.
With this experiment, if the null hypothesis is rejected, it means that the evaluated \ac{AE} method is better than the classical automatic-exposure method.
In the other case, since the Mann-Whitney \textit{U} test has already established that the distributions differ, failing to reject the null hypothesis implies that the evaluated distribution is inferior to the reference.
This evaluation technique enables the quantification of whether an approach has a significant impact relative to the defined metrics.

\section{EXPERIMENTAL SETUP}
\label{sec:experimental_setup}


This section outlines the experimental setup, beginning with a detailed description of our custom backpack for data acquisition.
We then describe the calibration procedure for the emulator and provide an overview of our large-scale multi-exposure dataset, \datasetName.
Finally, we discuss an extension of this dataset, where the same trajectory is repeated six times throughout the day to capture variations in illumination.

\subsection{DATA ACQUISITION BACKPACK}
\label{sec:experimental_setup_backpack}

Due to the challenging aspects of acquiring bracketing images at high frame rate and quality, we develop a custom data collection platform which gives us full control on the system.
In this section, as one of the main claims of this article is the need for more reproducible system, we provide an in-depth description of the standard components, such as the sensors, the main hardware, and the electrical connections.
In addition, we also analyze the performances of the platform and its limitations.
All the implemented software is available on GitHub.\footnote{\url{https://github.com/norlab-ulaval/backpack_workspace}}

\subsubsection{Overview}
\label{sec:experimental_setup_backpack_overview}

\begin{figure}[htbp]
    \centering
    \vspace{0.1in}
    \includegraphics[width=0.49\textwidth]{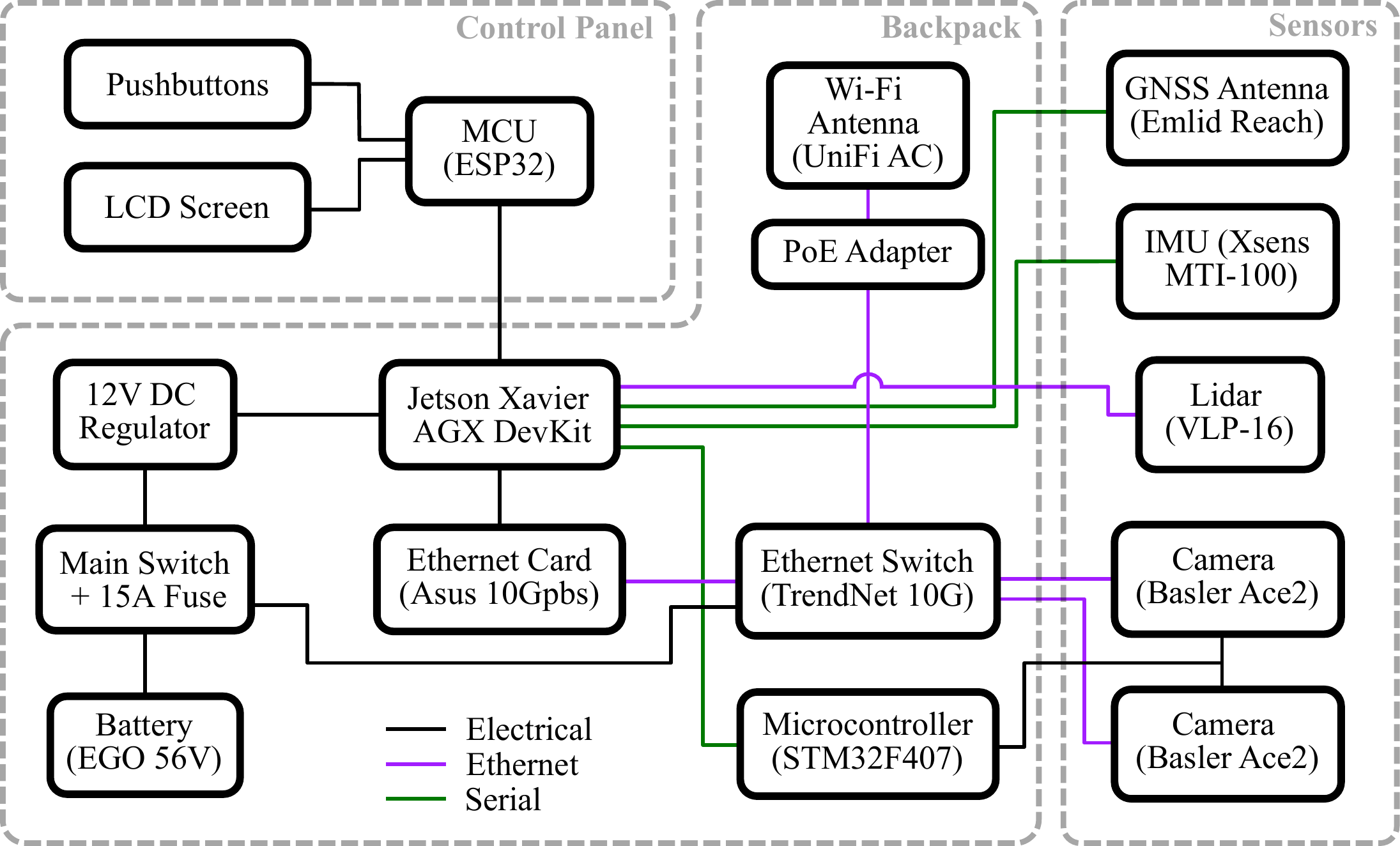}
    \caption{Block diagram of the hardware components of our backpack.
    Black lines show electrical connections, purple lines stand for Ethernet links and green lines are USB or serial cables.}
    \label{fig:schematic_hardware} 
\end{figure}

The platform is composed of three main sections: the backpack itself, the sensors, and the control panel.
A block diagram of the backpack's hardware components is presented in \autoref{fig:schematic_hardware}, where the subsystems and the links between them are displayed.
The figure also shows the interface type connecting each module, \ie~electrical, Ethernet, or serial.
Note that the final choice of the presented modules was determined after three version iterations of the platform, which were intertwined by outdoor testing and multiple deployments, described in \autoref{sec:lessons_learned}.

\subsubsection{Sensors}
\label{sec:experimental_setup_backpack_sensors}

\begin{figure}[htbp]
    \centering
    \includegraphics[width=0.47\textwidth]{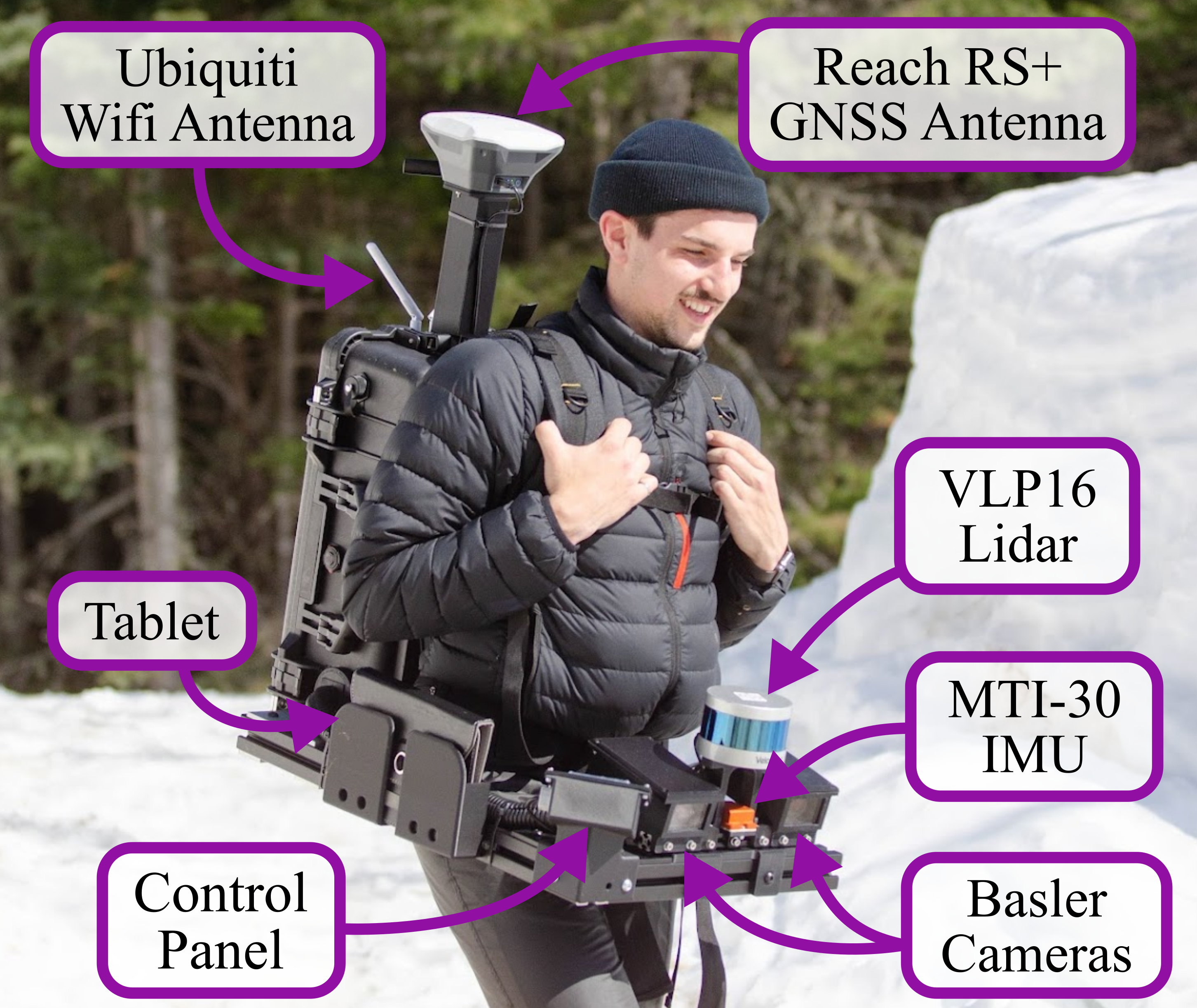}
    \caption{Picture of the developed backpack for the dataset acquisition. Main components are: Two Basler a2A1920-51gcPRO cameras in protective housing, a Xsens MTI-30 \ac{IMU}, a VLP16 3D lidar, an Emlid Reach RS+ \ac{GNSS} receiver, and an Ubiquiti UniFi UAP-AC-M wifi antenna.}
    \label{fig:backpack}
\end{figure}

The platform contains five main sensors as depicted in~\autoref{fig:backpack}.
For vision, we integrated two industrial cameras a2A1920-51gcPRO from Basler in a stereo-calibrated configuration with a baseline of \SI{18}{\centi\meter} and hardware-triggered system using the microcontroller mentioned above.
Each camera captures \num{12}-bit channel images using BayerRG encoding and has a \SI{1}{\giga b \per \second} bandwidth.
In addition to the camera, the platform is equipped with a Velodyne VLP-16 lidar, which also communicates using Ethernet.
The lidar is plugged into its own Ethernet card to mitigate any risk of entanglement between signals, leading to loss of information.
Finally, the platform is equipped with an Xsens MTI-30 \ac{IMU} and an Emlid Reach RS+ \ac{GNSS} receiver.
In the extended version of our dataset, we replace the GNSS antenna with an Emlid Reach M2 module, on which we receive \ac{RTK} corrections.
The corrections are sent by a base station from an Emlid Reach RS3+ communicating via a LORA radio module.
Similar to \citet{Jiajun2020}, the extrinsic calibrations between the sensors are given by the 3D model of the platform.
The intrinsic calibrations for the cameras were also calculated, by using a standard checkerboard.
The sensors all communicate through the \ac{ROS} Melodic middleware, which handles the timestamping of incoming data. 

\subsubsection{Hardware}
\label{sec:experimental_setup_backpack_backpack}

\begin{figure}[htbp]
    \centering
    \includegraphics[width=0.47\textwidth]{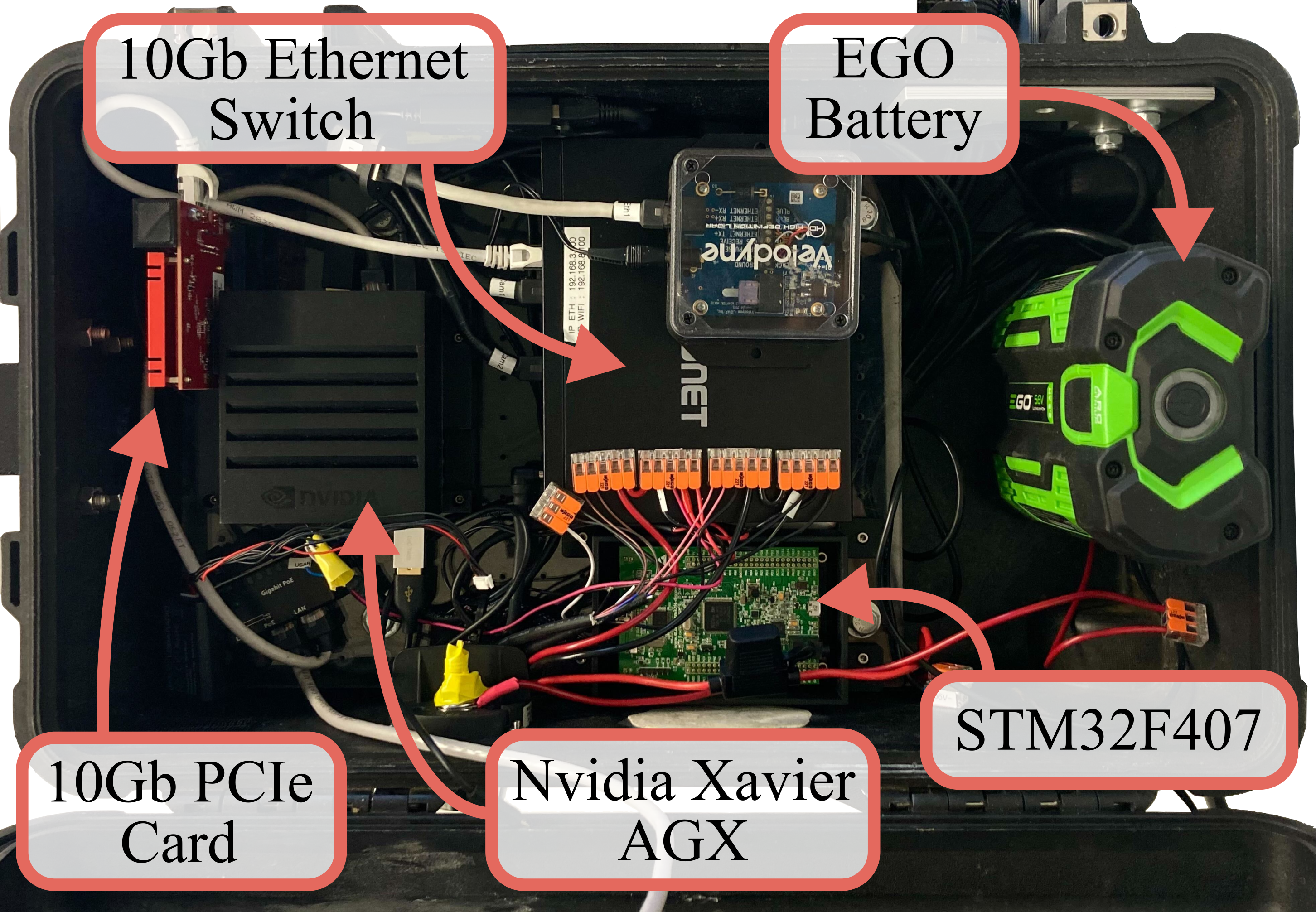}
    \caption{Picture of the inside of the backpack. Main components are: TRENDnet TEG-S762 switch, an \SI[detect-weight=true,mode=text]{2.5}{\ampere\hour} EGO battery, a STM32F407 microcontroller, a Nvidia Jetson Xavier AGX Developer Kit, and an Asus XG-C100C \SI[detect-weight=true,mode=text]{10}{\giga b\per\second} PCIe.}
    \vspace{-0.2in}
    \label{fig:backpack_inside}
\end{figure}

The backpack's core frame is a Pelican Case 1510, which is roughly the size of a person's back and has the advantage of being water-resistant.
The main power comes from a rechargeable \SI{56}{\volt} Ego battery installed on a custom-made 3D printed connector, depicted in \autoref{fig:backpack_inside}.
Different battery formats can be used, allowing to adjust the backpack's total weight between \SI{18}{\kilo\gram} and \SI{20.5}{\kilo\gram}.
The lightest battery is \SI{2.5}{\ampere\hour} and lasts around \SI{1.7}{\hour}, while the heaviest is \SI{7.5}{\ampere\hour} for a total battery life of \SI{5.5}{\hour}.
The embedded computer is a Jetson Xavier AGX Developer Kit, which records all the sensors' data directly into a 970 EVO Plus NVMe M.2 SSD of \SI{1}{\tera\byte}.
This computing unit is equipped with two embedded \SI{1}{\giga b\per\second} Ethernet interfaces.
Resulting from a lack of bandwidth that led to data losses, the computer was complemented with an Asus XG-C100C \SI{10}{\giga b\per\second} PCIe card that is connected to a TRENDnet TEG-S762 switch, offering two \SI{10}{\giga b\per\second} ports and four \SI{2.5}{\giga b\per\second} ports.
For better control over the image acquisition, an STM32F407 microcontroller acts as an external trigger to synchronize the acquisition from both cameras, but also for precise exposure time control.
The trigger signal is generated using hardware timers in the microcontroller, providing more robust timing than software-based solutions.
Note that we tried the integrated sequencer feature in the industrial cameras, but it led to inconsistent frame grabbing, hence the need for an external microcontroller. 
Finally, an antenna from Ubiquiti, the UAP-AC-M UniFi AC Mesh AP,  is connected to its own Ethernet card, and serves as a hotspot to allow wireless connection into the backpack's computer.

\subsubsection{Control Panel}
\label{sec:experimental_setup_backpack_panel}

To improve user awareness of the system state, we add an ESP32 microcontroller that communicates with the Jetson Xavier AGX over USB to display the status of each sensor to the user on an LCD screen in front of the platform.
It shows five types of messages:
\begin{itemize}
    \item \texttt{OFF}: The sensor is not used.
    \item \texttt{IDLE}: The sensor is ready for data acquisition.
    \item \texttt{REC}: The sensor's data is being recorded.
    \item \texttt{ERR}: There is a problem with the sensor.
    \item \texttt{CAL}: The sensor is undergoing calibration (\ac{IMU} only).
\end{itemize}
Finally, the screen also shows the remaining free space on the drive and the cameras' acquisition rate.
The ESP32 is also connected to two push buttons, which allow to start the sensors and to enable the data recording.
We integrate a compact and lightweight visualization tablet mounted on the side of the backpack, enabling efficient debugging of camera images and lidar scans.


\subsubsection{Performances Analysis}
\label{sec:experimental_setup_backpack_performances}


Based on the components installed in the platform, we analyzed the necessary bandwidth while recording data from all the sensors.
The Ethernet transmission rate is examined for each of the three Ethernet cards installed on our Jetson Xavier AGX, \ie~the two embedded interfaces and the external PCIe card.
The bandwidth coming from the camera is the highest with around \SI{872}{\mega \bit/\second}, while the transmission rate from the lidar is around \SI{8}{\mega \bit/\second}.
The last Ethernet card is plugged directly into the Ubiquiti antenna, which uses no bandwidth unless another computer is connected to it for visualization. 
An analysis of the total CPU load showed that an average of \num{5.5} cores are used from the \num{8} available, while the system uses \SI{1.55}{\giga\byte} of RAM during recording.

In relation to the performances, we estimated the energy consumption needed to power all the hardware in the platform.
The total consumption is approximately \SI{79}{\watt}, and is mainly from the computer and the Ethernet switch, using respectively \SI{40}{\watt} and \SI{12}{\watt}.

\subsection{EMULATOR CALIBRATION SEQUENCES}
\label{sec:experimental_setup_static}

To evaluate the performance of our image emulator, it was necessary to acquire images with known exposure times to serve as ground truth for comparison. 
For this purpose, we recorded sequences in static scenes without moving the platform, capturing images at multiple exposure times. 
In total, five sequences comprising \num{1000} ground truth images were collected, with exposure times ranging from \SI{20}{\micro\second} to \SI{50}{\milli\second}. 
This dataset includes images from both indoor and outdoor static scenes. 
Additionally, to account for the camera's intrinsic noise, we captured \num{25} identical images at several exposure times in the spectrum. 
By averaging these images, the camera's noise characteristics across the entire exposure time range can be estimated.
Furthermore, these calibration sequences were used to compute the \acf{CRF} as described in~\autoref{sec:theory_emulation}, following the method proposed by~\citet{engel2016monodataset}.

\subsection{\datasetName~DATASET}
\label{sec:experimental_setup_dynamic}

As one of the contributions consist of analyzing the impact of \ac{AE} methods in the field, we present our extended \datasetName~dataset recorded to support our emulator and provide reference modalities for comparative analysis.
Our \datasetName~dataset, and its new extension, depicted in \autoref{fig:borealhdr}, comprises \num{\numberTrajectories}~sequences totaling around \kilometerTravelled~\si{\kilo\meter} for \totalHours~\si{\hour}, and around \SI{3.5}{\tera\byte} of data.

\subsubsection{\datasetName}
\label{sec:experimental_setup_dynamic_borealhrd}

As described in \autoref{sec:experimental_setup_backpack}, we gather our data using our custom backpack.
The images, which are of dimension $1920 \times 1200$, are acquired at a rate $r_\text{real}$ of \num{22}~\ac{FPS} in \num{12}-bit color, using lossless compression.
From the previously acquired version of \datasetName~\citep{gamache2024exposing}, the image acquisition process cycles through $N=\si{6}$ exposure time values, \ie{} $\BracketList=\{1,2,4,8,16,32\}$~\si{\milli\second}, yielding an effective emulation rate of $r_\text{emul}=$ 3.66 \ac{FPS}.
This number of brackets is a compromise between the offline $r_\text{emul}$ and the emulation error, detailed in~\autoref{sec:results_emulation}.
The $\BracketList$ values were chosen to cover the whole dynamic range of our target environments. 
The exposure time is doubled at every step of the cycle, which is equivalent to a one-stop increase in photography.
We also include data from other modalities, with the 3D lidar collecting point clouds at \SI{10}{\hertz}, the \ac{IMU} at \SI{100}{\hertz}, and the \ac{GNSS} receiving messages at \SI{1}{\hertz}.

Half of the trajectories are loops, which is a common practice to estimate drift in visual \ac{SLAM} datasets.
The relatively low emulation rate $r_\text{emul}$ implies that the data should be collected at a slow walking speed to minimize displacements between each acquisition cycle.
Hence, our average pace is around \SI{2}{\kilo\meter/\hour}, which increases the data collection time but does not impact the users of our offline emulation pipeline.
In total, \numberImages~images were gathered on four separate days in summer and winter \num{2023}.
The winter data, from April 20$^{\text{th}}$ to April 21$^{\text{st}}$, was acquired in the Montmorency \textbf{Forest} in Canada.
The summer trajectories were collected in the Québec City area, in Mont-\textbf{Bélair} natural park and on the \textbf{Campus} of Université Laval, at the end of September.
There are \num{42} trajectories from the \textbf{Forest} subset, \num{8} from \textbf{Bélair} and \num{3} acquired on the \textbf{Campus}, each site respectively totaled \SI{7.2}{\kilo\meter}, \SI{2.2}{\kilo\meter}, and \SI{0.6}{\kilo\meter}.
The resulting data from \textbf{Forest} strongly emphasizes highly contrasted \ac{HDR} scenes, by capturing the dark trees and the bright snow altogether, visible in the top part of \autoref{fig:borealhdr}.
Meanwhile, the summer data from \textbf{Bélair} and \textbf{Campus} targeted trajectories with a lot of illumination variations, alternating between sunlight and shadowy areas. 
This contrast is clearly seen by looking at individual rows in \autoref{fig:borealhdr}, showcasing that scene irradiance can vary significantly over a single trajectory.  
In addition, multiple trajectories contain challenging conditions for vision-based localization, including sun glares and texture-less environments such as immaculate snow or a lake's steady water.
In winter, we used snowshoes to get to remote locations, to collect not only different scene illuminations, but also various 3D structures, allowing for a richer collection of test scenarios. 
The small footprint of the backpack allows for collecting data in narrow and hard-to-access spaces for robotic vehicles, which was fundamental for our \kilometerTravelled~\si{\kilo\meter} dataset. 

\subsubsection{\datasetName~Extension}
\label{sec:experimental_setup_dynamic_extension}

\begin{figure*}[htbp]
    \centering
    \vspace{0.1in}
    \includegraphics[width=0.98\textwidth]{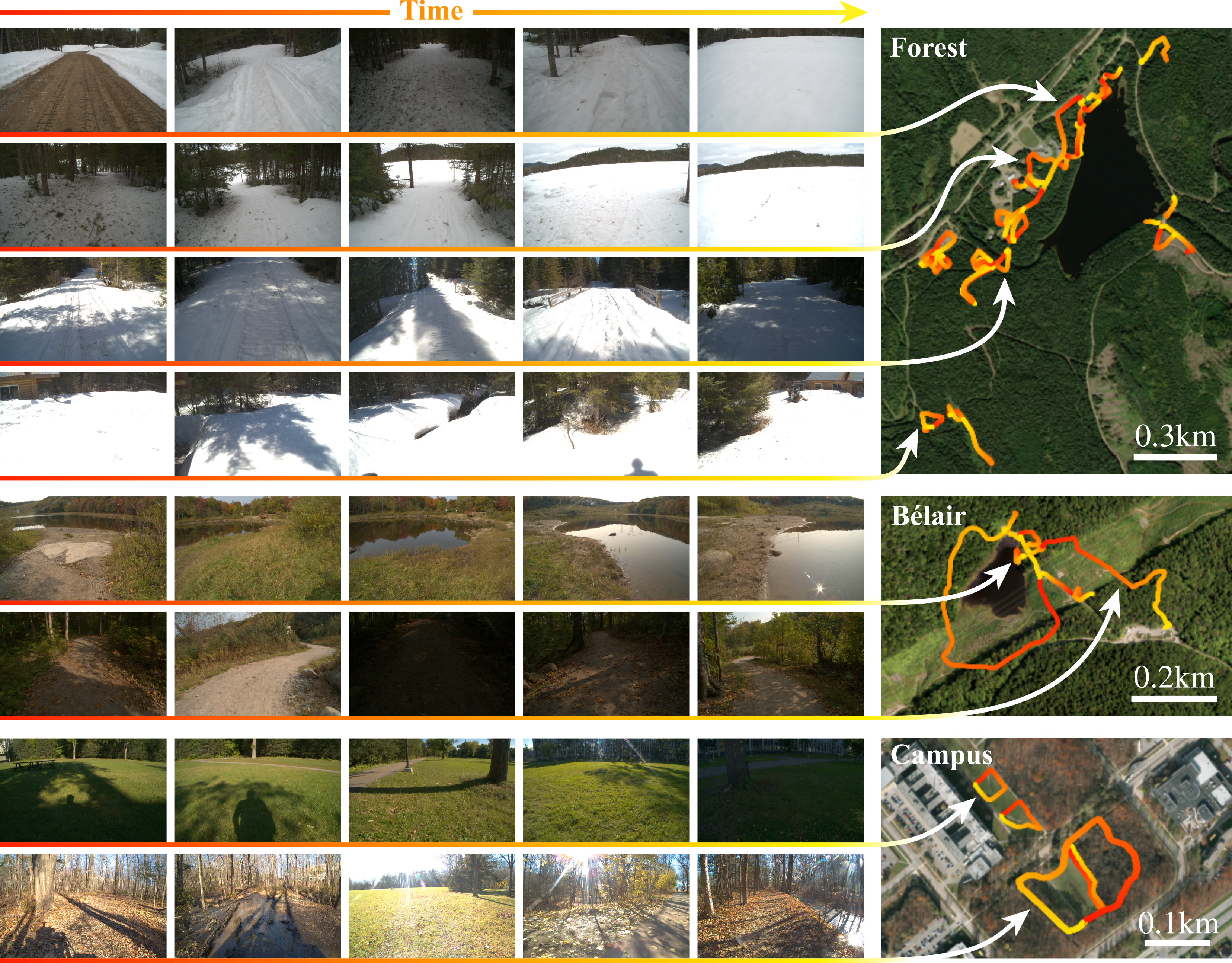}
    \caption{Overview of the BorealHDR dataset, composed of \num[detect-weight=true,mode=text]{\numberTrajectories}~trajectories in three different environments, hereby named Forest, Bélair and Campus. The left part shows samples from selected trajectories, emphasizing the diversity of the collected data, especially in terms of scene's dynamic range and illumination variations. 
    The red to yellow gradient represents the relative progression in a trajectory.
    The right part highlights the wide areas covered for the data collection over satellite images, obtained through \textcopyright OpenStreetMap. 
    Note that all the displayed images are using a fixed exposure time of \SI[detect-weight=true,mode=text]{8}{\milli\second}, except for the last row where the aperture changed, hence the selected exposure time is \SI[detect-weight=true,mode=text]{1.6}{\milli\second}.
    }
    \label{fig:borealhdr}
\end{figure*}

This paper brings forth an extension to the previously described \datasetName~dataset. 
Unlike the original version, where a trajectory was captured under a single illumination condition, this extension involves capturing the same trajectory six times throughout a single day.
The backpack was enhanced by increasing the lenses’ aperture, reducing motion blur at higher exposure times in the bracketing cycle.
The updated exposure time values are $\BracketListExtension=\{0.025,0.1,0.4,1.6,6.4,25.6\}$~\si{\milli\second}.
The fourfold increase in exposure time between successive images allows for extended dynamic range coverage.
Additionally, as detailed in~\autoref{sec:experimental_setup_backpack_sensors}, \ac{RTK}-\ac{GNSS} data were recorded for each trajectory.
Image samples from this repeated trajectory are shown in the last row of \autoref{fig:borealhdr}.

To accurately record the same trajectory multiple times, the backpack was mounted on a Clearpath Warthog platform to ensure a consistent displacement speed. 
The \ac{WILN} Teach-and-Repeat framework,\footnote{\url{https://github.com/norlab-ulaval/wiln}}, developed by \citet{Baril2022}, was employed to achieve precise trajectory replication. 
The teach trajectory was recorded in the morning of November 13$^{\text{th}}$ 2024, on Université Laval’s campus, with subsequent recordings repeated every two hours from 7:50 AM until dusk at approximately 4:05 PM.
Following the original version of the dataset, the repeat speed was maintained at a constant \SI{2}{\kilo\meter/\hour}.
The trajectory is a loop of approximately \SI{570}{\meter}, which primarily traverses open sky areas and forest trails. 
By capturing images throughout the day, we incorporated diverse illumination conditions, including sun glare, shadows, and sun-free skies at dusk.
Note that for the remainder of the paper, this extension will be referred to as \textbf{Campus2}, and the original campus subset as \textbf{Campus1}.

\subsubsection{Reference Trajectories}
\label{sec:experimental_setup_dynamic_reference}

In addition to the cameras, the backpack is equipped with a lidar and a \ac{GNSS} antenna, which both provide useful information to estimate the traveled trajectories.
In boreal forests,~\citet{Baril2022} demonstrated that \ac{GNSS} signals are unreliable, due to attenuation caused by the dense vegetation canopy.
Therefore, our reference trajectories were generated by employing a low-drifting lidar-inertial-odometry variant~\citep{Kubelka2022}, based on the libpointmatcher library.\footnote{\url{https://github.com/norlab-ulaval/libpointmatcher}}
This method registers deskewed point clouds, at a rate of \SI{10}{\hertz}, into a dense 3D map of the environment.
It relies on the on-board lidar and is illumination-independent, with median error values around \SI{1}{\percent} of the total trajectory length.
The 3D maps were computed in post-processing to optimize the accuracy and overall quality.
Note that, we do not assume that the reference trajectories are equivalent to the ground truths, but that they provide adequate precision for comparison with the visual trajectories, as explained in~\autoref{sec:results_benchmark_vo}.

\subsection{IMPLEMENTATION DETAILS OF COMPARED \ac{AE} ALGORITHMS}
\label{sec:experimental_setup_implementation_details}

Based on~\autoref{sec:related_work_ae}, we implement and benchmark four state-of-the-art \ac{AE} methods: \textbf{Shim}~\citep{Shim2019}, \textbf{Wang}~\citep{Wang2022}, \textbf{Kim}~\citep{Kim2020}, and \textbf{DRL}~\citep{zhang2024efficient}. 
The methods from \textbf{Shim} and \textbf{Kim} are not open-source, thus they were implemented based on our understanding of the papers.
\textbf{Wang} relies on a heuristic search-based exposure prediction algorithm, where images are emulated in a separate thread to estimate the optimal exposure time for the next frame.
In our implementation, this algorithm is computed $n=3$ time before determining the exposure time for the subsequent image.
The exposure control scheme of \textbf{Kim} used a Gaussian process, which sparsely sweep the camera exposure parameters until convergence.
In our implementation, we use a sliding window on the trajectory to only consider the most recent images in the Gaussian process training.
This can cause steep changes in the desired exposure time, since the exploration algorithm of the Gaussian process needs to cover a wide range of values to converge.
For \textbf{DRL}, we directly include the deep reinforcement agent that was trained on their feature reward directly on their dataset.
To evaluate the domain adaptation, we did not fine-tune their agent on our dataset.
We also implemented four baseline methods, corresponding to the typical \ac{AE} algorithms.
The first baseline algorithm is a fixed exposure time approach, which we called \textbf{Fix}.
This emulated exposure time is selected once, at the beginning of each sequence, by using a brightness target of \SI{50}{\percent}.
The three other baselines are variable exposure algorithms, seeking to keep the average brightness target to \SI{30}{\percent}, \SI{50}{\percent}, and \SI{70}{\percent} of the \num{12}-bit depth range. 
They are named respectively \textbf{AE30}, \textbf{AE50} and \textbf{AE70}.
These percentages were chosen as to cover a wide range of exposure.
Note that our implementation of all the described methods is available in our GitHub repository.

\section{RESULTS}
\label{sec:results}

\begin{figure*}[htbp]
	\centering
        \vspace{0.1in}
	\includegraphics[width=0.98\textwidth]{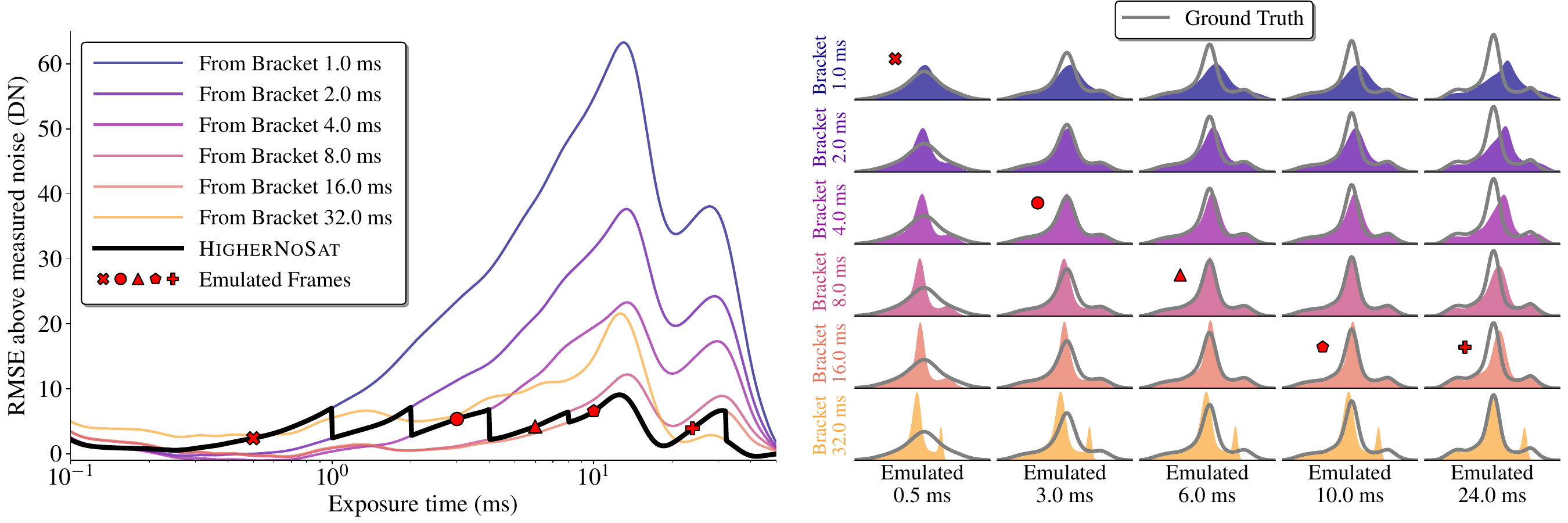}
	\caption{Validation of our emulation framework using \num[detect-weight=true,mode=text]{1000} ground truth images captured with constant illumination, but varying exposure time. \textit{\textbf{Left:}} \ac{RMSE} curves showing the emulation error if a single bracket was always selected, in colors, compared to our bracket selection method \selectionMethod, in black. Red symbols correspond to the emulated exposure times displayed on the right. \textit{\textbf{Right:}} Qualitative comparison of the distributions of five emulated exposure times (columns) from six different bracket images (rows). The markers are placed next to the selected bracket by \selectionMethod. Ground truth distributions are overlaid in gray for each column.
            }
	\label{fig:emulation_performances}
\end{figure*}

This section provides an analysis of our emulator's performance and outlines our replicable and scalable benchmarking methodology.
Using the metrics defined in~\autoref{sec:theory_experiments}, we evaluate whether state-of-the-art \ac{AE} methods demonstrate a significant impact on visual \ac{SLAM} performances when applied in diverse outdoor environments.

\subsection{EMULATION STATIC VALIDATION}
\label{sec:results_emulation}

Considering that the upcoming experiments are based on our emulation pipeline, we first establish the photographic realism of the emulated images.
As described in~\autoref{sec:experimental_setup_static}, the performance of our image emulator is evaluated using five test sequences, each comprising \num{1000} ground truth images captured at exposure times ranging from \SI{20}{\micro\second} to \SI{50}{\milli\second} in both indoor and outdoor static scenes.
Using these images as ground truth $I_\text{GT}$, we emulate images $I_\text{emul}$ with the same exposure times as the ground truths.
To account for the intrinsic noise of the camera, which varies with exposure time, we average the difference between \num{25} images from the same exposure time.
This process is repeated across multiple exposure times, enabling interpolation of the camera noise over the entire exposure spectrum.
The \acf{RMSE} between $I_\text{emul}$ and $I_\text{GT}$, adjusted for camera noise, is then calculated to assess the emulator's capability to produce realistic images.
The five test sequences serve to validate our bracket selection method \selectionMethod, described in \autoref{alg:pseudo_code}, which does not have access to ground truth images when deployed in real world settings.
Across all five sequences, \selectionMethod~achieves a median \acf{RMSE} of \num{9}~\ac{DN} and never surpasses \num{73}~\ac{DN}.
For \num{12}-bit image, these values correspond respectively to \SI{0.21}{\percent} and \SI{1.78}{\percent} of relative error, highlighting the method's consistency and accuracy.
Note that for this experiment, we use $N=\si{6}$ exposure time values being $\BracketList=\{1,2,4,8,16,32\}$~\si{\milli\second}, and a saturation threshold of $\alpha = 0.01$ for the emulator.

One of the test sequences is detailed in~\autoref{fig:emulation_performances}, showing emulation performances in controlled settings.
On the left side, the \ac{RMSE} curve from \selectionMethod, in black, is compared to the error obtained by selecting a single bracket as $I_\text{source}$ for the whole range, in colors.
It shows that our bracket selection maintains the error close to the lowest value of the \numberExposureBracketing~colored curves.
It is important to note that the method is optimized to achieve robust average performance across most cases, as ground truth images are not available during real data acquisition.
Consequently, the \selectionMethod~curve may not always align with the closest possible value.
The right side plots present a qualitative evaluation of the emulated images from the acquired brackets.
Each column exposes the distributions of pixel intensity for one target exposure time, while each row emulates this image from a different bracket.
Red markers represent which exposure time in $\BracketList$ was chosen by our emulator via \selectionMethod, for this image.
We observe that our bracket selector consistently picks the closest or second-closest emulated image to the ground truth distribution, overlaid in gray.

\subsection{PERFORMANCE ANALYSIS OF THE \ac{AE} METHODS}
\label{sec:results_benchmark}


Benefiting from our unique dataset, one of the main contribution of our paper is to unlock \emph{offline} testing of active vision methods that could only previously be tested in an \emph{online} manner. 
Using our setup described in~\autoref{sec:experimental_setup}, we analyze the performance of the implemented \ac{AE} methods on \datasetName~and its new extension.
To this effect, we conduct several experiments to highlight the impact of each \ac{AE} method in key aspects of a visual \ac{SLAM} pipeline, such as the feature distribution, the \ac{RPE}, and their time before failure.
By the end of this section, we conclude how the state-of-the-art \ac{AE} methods perform against the classical automatic-exposure control method \textbf{AE50}.

We conduct the subsequent experiments using a slightly modified version of stereo ORB-SLAM2~\citep{mur2017orb} as our visual \ac{SLAM} algorithm.
Due to the low image rate $r_\text{emul}$ and the roll and pitch oscillations introduced by data acquisition using the backpack, we noticed that ORB-SLAM2 have some difficulties to converge when relying on its motion model as a prior.
To enhance its performance, we disable the motion model and configure the system to use all images as keyframes.
While this adaptation makes the modified ORB-SLAM2 unsuitable for real-time applications, it is well-suited for offline testing.
Additionally, we disable loop closure detection to ensure a fair comparison of the \ac{AE} methods.
The ORB-SLAM2 parameters are set to \textsc{nFeatures}~=~\num{3000}, \textsc{scaleFactor}~=~\num{1.2}, \textsc{nLevels}~=~\num{13}, \textsc{iniThFAST}~=~\num{15}, and \textsc{minThFAST}~=~\num{5}.

\subsubsection{Feature Tracking and Uniformity}
\label{sec:results_benchmark_features}

\begin{figure}[htbp]
    \vspace{3mm}
    \centering
    \includegraphics[width=0.47\textwidth]{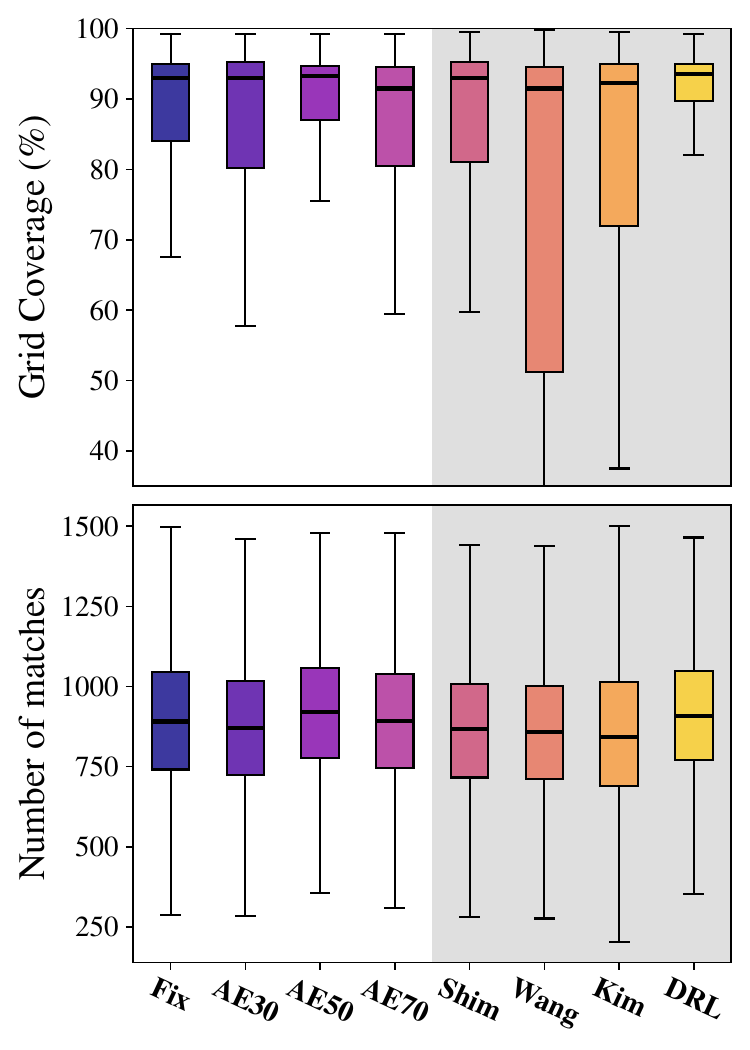}
    \caption{ORB-SLAM2 features analysis. (Top) Uniformity of detected keypoints for an image divided into a \bm{$20\times20$} grid. (Bottom) Number of match features with the visual 3D map for all \ac{AE} metrics.
    The gray shading is to highlight the state-of-the-art \ac{AE} methods.}
    \label{fig:keypoints}
\end{figure}

The performances of feature-based visual \ac{SLAM} algorithms such as ORB-SLAM2 is highly dependent on the quality and quantity of detected keypoints.
Therefore, we evaluate feature detection for each implemented \ac{AE} algorithm on our datasets, over two distinct tests.
The first test consists of estimating the uniformity of the detected keypoints in the image, by dividing each image into a $20\times20$ grid and assessing its occupancy rate.
The results are displayed in coverage percentage in the top graph of~\autoref{fig:keypoints}, which shows that the implemented \ac{AE} methods yield a similar grid coverage.
This effect is mostly due to ORB-SLAM2, which already divides the input image into a grid for the keypoints search.
Nevertheless, we observe that the two most effective methods are \textbf{AE50} and \textbf{DRL} with a median value of \SI{93.5}{\percent} and \SI{93.0}{\percent} respectively, while the least performing methods in this experiment are \textbf{AE70} and \textbf{Wang} with a median of \SI{91.5}{\percent}.
Interestingly, we notice that these last two methods are also the ones that tend to request higher exposure times.
Since more than \SI{70}{\percent} of the trajectories contain snow, the high exposure times saturate the pixels corresponding to the snow, making keypoints detection challenging.

The second test consists of comparing the number of matched keypoints between the current frame and ORB-SLAM2's computed map, for each image. 
The results' distributions are presented in the bottom graph of~\autoref{fig:keypoints}, for each of the implemented \ac{AE} methods. 
In accordance with the coverage experiment, the top-performing methods are \textbf{AE50}, with a median of \num{920} matches, and \textbf{DRL}, with \num{908} matches. 
The strong performance of the \textbf{DRL} method in both feature-based experiments is to be expected, as its reward function, described by \autoref{eq:reward_fct}, seeks to maximize both the number of detected features and the number of matches between consecutive images.
These results confirm that the \textbf{DRL} method improves feature quality and quantity, even when applied to a different dataset than the one used for training.
Interestingly, despite its simplicity, the \textbf{Fix} method also performs well, achieving a median of \num{891} matches. 
This can presumably be attributed to the absence of exposure changes, which maintains a smoother flow between consecutive images. 
These findings suggest that minimizing abrupt variations in exposure levels positively impacts the feature-matching process.

\subsubsection{Relative Trajectory Error}
\label{sec:results_benchmark_vo}

\begin{figure}[htbp]
	\centering
	\includegraphics[width=0.48\textwidth]{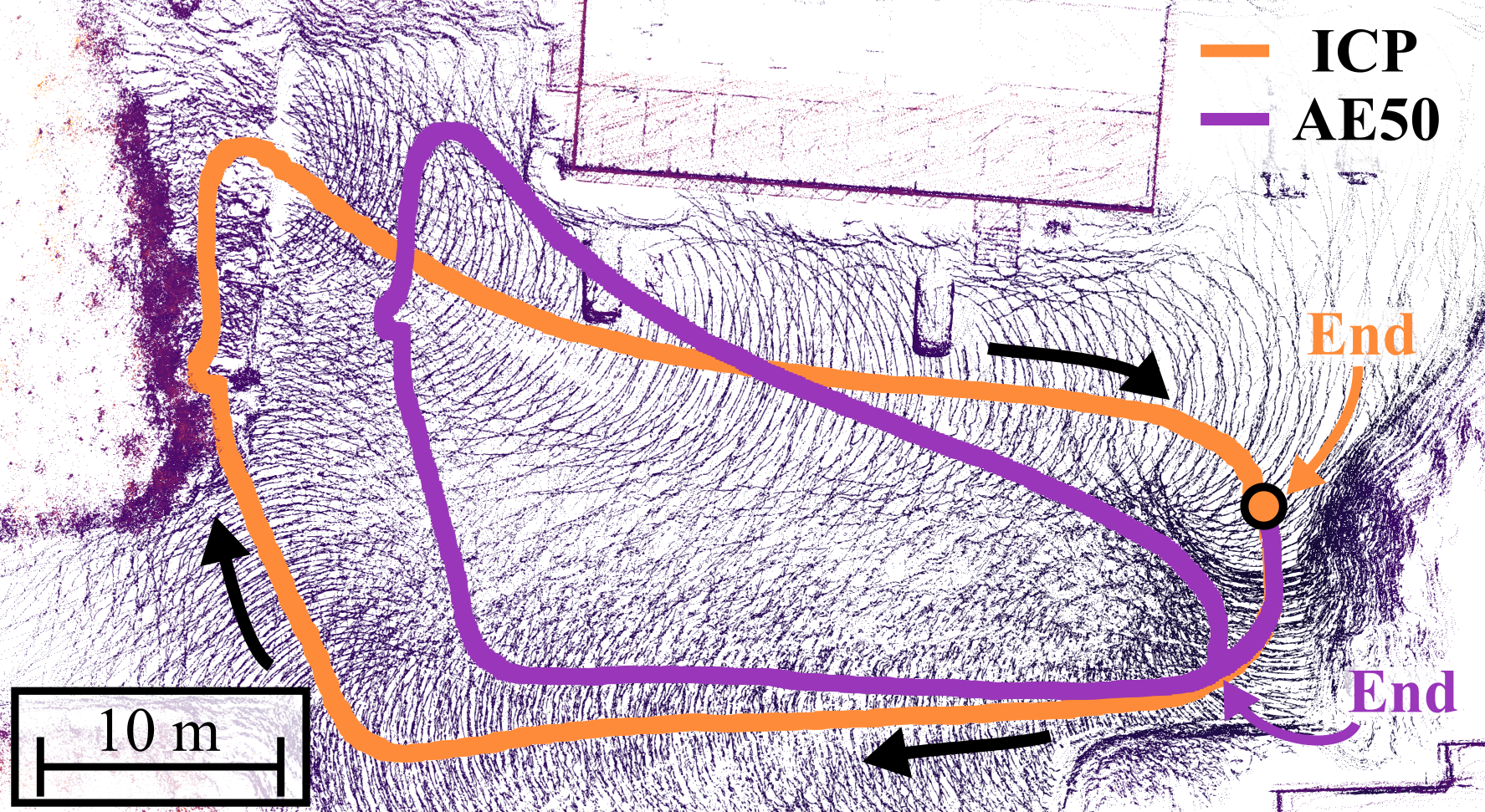}
	\caption{
            Demonstration of the lidar-based localization and mapping used as a reference for the visual \ac{SLAM} evaluation.  
            This example illustrates the trajectory \texttt{backpack\_2023-04-21-08-51-27} from our \datasetName~dataset, where the start and end points are the same.
            The reference trajectory (orange) and \textbf{AE50}'s emulated ORB-SLAM2 result (purple) are superimposed on the generated lidar 3D~map. 
            }
	\label{fig:results_lidar_gt}
\end{figure}

We also investigate the impact of \ac{AE} methods on the ORB-SLAM2 visual \ac{SLAM} pipeline.
To evaluate the performances of each method, we compare the output trajectories from ORB-SLAM2 to our lidar-inertial-odometry reference trajectories described in~\autoref{sec:experimental_setup_dynamic_reference}.
An example using one trajectory from our dataset is illustrated in~\autoref{fig:results_lidar_gt}, along with the 3D map.
The observed offset between the reference trajectory and ORB-SLAM2 is mainly due to the accumulated drift in elevation, common to the visual \ac{SLAM} method and usually mitigated by the loop closure \citep{mur2017orb}, which was disabled in the spirit of equitable benchmarking.

As described in~\autoref{sec:theory_experiments_vslam}, we evaluate the performance of each \ac{AE} method on ORB-SLAM2 by computing the \ac{RTE} \eqref{eq:RTE} and the \ac{RRE} \eqref{eq:rre}.
We use the \emph{evo}\footnote{\url{https://github.com/MichaelGrupp/evo}} library to compute the relative errors, with the set of evaluation windows $\bm{W}$ containing every integer between \num{5} and \num{50}, inclusively.
The final results, summarized in~\autoref{table:rpe}, demonstrate that the predictive \textbf{DRL} method consistently achieves the lowest localization errors. 
The classical illumination-based method, \textbf{AE50}, closely follows as the second-best performer. 
Interestingly, all state-of-the-art reactive methods exhibit lower accuracy in trajectory estimation compared to \textbf{AE50}.
This discrepancy can be explained from the fact that reactive methods often require extensive parameter tuning and are prone to overfitting to specific scenarios. 
In contrast, the classical \textbf{AE50} method is more general and exhibits greater robustness across our diverse dataset. 
Predictive methods like \textbf{DRL}, however, leverage the temporal context by using recent frames to predict future illumination changes, offering a distinct advantage over reactive methods that rely solely on the current frame.


\begin{table}
    \setlength{\tabcolsep}{3.5pt}
    \renewcommand{\arraystretch}{1.2}
    \caption{Relative pose error between ORB-SLAM2 trajectories and our reference lidar-based trajectories.}
    \begin{tabular}{L{1.4cm}|cccccccc}
        \hline
        Metric & \textbf{Fix} & \textbf{AE30} & \textbf{AE50} & \textbf{AE70} & \textbf{Shim} & \textbf{Wang} & \textbf{Kim} & \textbf{DRL} \\
        \hline
        \ac{RTE} (\%) & 19.5 & 20.8 & 18.8 & 24.7 & 22.6 & 32.1 & 27.6 & \textbf{18.4} \\
        \ac{RRE} (\si{\degree/\meter}) & 0.37 & 0.40 & 0.34 & 0.35 & 0.39 & 0.44 & 0.36 & \textbf{0.33} \\
        \hline
    \end{tabular}
\label{table:rpe}
\end{table}


\subsubsection{Robustness Evaluation}
\label{sec:results_benchmark_robustness}

\begin{figure}[htbp]
    \vspace{3mm}
    \centering
    \includegraphics[width=0.47\textwidth]{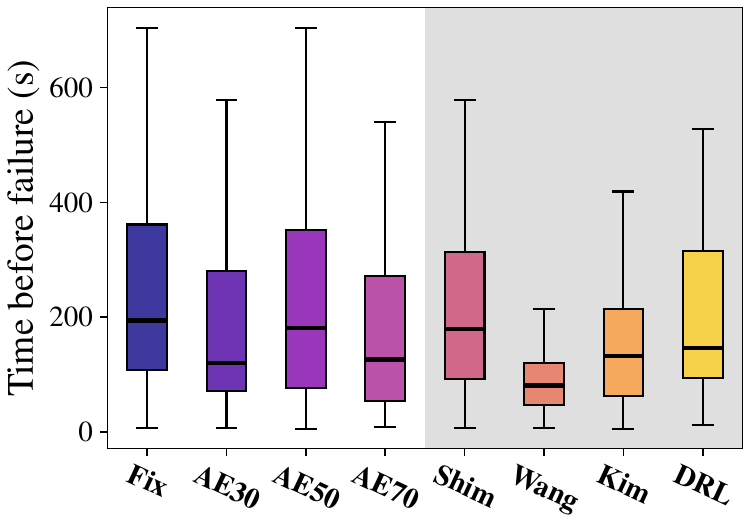}
    \caption{The trajectory duration without having a first failure using ORB-SLAM2.
    The gray shading is to highlight the state-of-the-art \ac{AE} methods.}
    \label{fig:failure}
\end{figure}

A critical and desirable attribute of visual \ac{SLAM} algorithms is their robustness in challenging environments.
To evaluate this characteristic, we first look at the time before failure for all the trajectories from our \datasetName~dataset and combine the results for each \ac{AE} method.
\autoref{fig:failure} highlights the results, where we see that \textbf{AE50} and \textbf{Fix} are the best performing algorithms. 
A closer examination of the failure cases reveals that failures commonly occur when there are abrupt changes in exposure time, making it difficult to match features with existing map points.
This issue is absent in \textbf{Fix}, as it maintains constant exposure, and transitions remain smooth with \textbf{AE50} due to its limited update rate.
On the other hand, methods like \textbf{Kim} incorporate exploration steps that cause significant exposure time variations, while \textbf{Wang}'s heuristic search-based exposure prediction algorithm can result in rapid exposure changes.
\textbf{Shim} demonstrates robustness against failures by ensuring smooth transitions through a slower convergence rate and tends to favor lower exposure times, as seen in \autoref{fig:results_exposure_comparison}.

\begin{table}
    \setlength{\tabcolsep}{3.5pt}
    \renewcommand{\arraystretch}{1.2}
    \caption{Number of successful trajectories using ORBSLAM2 for each deployment day.
    The total number of trajectories for each day is written in parentheses in the first column.}
    \begin{tabular}{L{1.6cm}|cccccccc}
        \hline
        \textbf{Day} & \textbf{Fix} & \textbf{AE30} & \textbf{AE50} & \textbf{AE70} & \textbf{Shim} & \textbf{Wang} & \textbf{Kim} & \textbf{DRL} \\
        \hline
        \textbf{Forest} \textbf{(42)} & 32 & 22 & 26 & 17 & 29 & 6 & 15 & 21 \\
        \textbf{Bélair} \textbf{(8)} & 4 & 3 & 3 & 2 & 5 & 2 & 3 & 3 \\
        \textbf{Campus1} \textbf{(3)} & 0 & 0 & 0 & 0 & 0 & 0 & 0 & 0 \\
        \textbf{Campus2} \textbf{(6)} & 4 & 4 & 2 & 4 & 5 & 4 & 1 & 3 \\
        \hline
        \textbf{Total} & 40 & 29 & 31 & 23 & 39 & 12 & 19 & 27 \\
        \hline
    \end{tabular}
\label{table:success_trajs}
\end{table} 

Continuing with the robustness analysis of each \ac{AE} method, we compiled the number of successful trajectories for each deployment day, as shown in~\autoref{table:success_trajs}. 
A successful trajectory is defined as one where ORB-SLAM2 remains consistently localized throughout the sequence without getting lost.
As anticipated, the results align with the trends observed in~\autoref{fig:failure}, with only \textbf{Fix}, \textbf{Shim}, and \textbf{AE50} managing to complete more than half of the dataset’s trajectories without a single failure. However, \textbf{DRL} underperforms compared to expectations, given its earlier success in feature tracking and relative trajectory errors.
Training \textbf{DRL} directly on our dataset could mitigate domain shift and improve robustness, as the model was not originally trained on natural environments.
\textbf{Kim} often fails due to exploratory phases that cause abrupt changes in exposure time, while \textbf{Wang} tends to converge to high exposure times, leading to saturation.
Because \textbf{Wang} simulates alternate exposure times based on the current frame, the algorithm struggles to lower exposure back to acceptable levels once saturated.
Both methods occasionally exhibit oscillatory behavior, likely a result of poorly tuned parameters.
This highlights their lack of generalization and the challenges of optimizing them for diverse scenarios.

On average, the tested methods successfully complete \num{27.5} trajectories, accounting for only \SI{47}{\percent} of the \datasetName~dataset, leaving significant room for improvement. 
The \textbf{Campus1} subset proved particularly challenging, with a lot of sun glare, resulting in failures for all trajectories.
Note that these results should be interpreted with caution, as ORB-SLAM2's non-deterministic nature can result in random failures. 
Additionally, the dataset includes motion blur at the highest exposure brackets, favoring \ac{AE} methods that prioritize lower exposure times.
In general, certain trajectories remain inherently difficult for visual \ac{SLAM} systems, with challenges unrelated to exposure control, such as abrupt camera movements, vegetation striking the lens, and texture-less environments.

\subsubsection{Statistical Analysis of Automatic-Exposure Methods}
\label{sec:results_benchmark_statistical}

To support our second contribution, we leverage the accumulated data and computed metrics from the preceding experiments to evaluate the real-world performance of state-of-the-art \ac{AE} methods in field deployments. 
Employing the Mann-Whitney \textit{U} test with a \SI{95}{\percent} confidence interval, as described in \autoref{sec:theory_experiments_statistical}, we conduct pairwise comparisons between each benchmarked \ac{AE} method and \textbf{AE50}, thereby determining whether any method demonstrates statistically significant improvement or deterioration in performance. 
For each trajectory in the dataset, we compute the mean values of the following metrics: feature coverage, number of matches, \ac{RTE}, \ac{RRE}, and time before failure.
By analyzing the distribution across the entire dataset, we obtain the results presented in \autoref{table:p_value_flipped}.

As indicated by the initial experiments, the \textbf{AE50} method performs equally or better than all other reactive methods, particularly for the relative error metrics. 
Its simplicity, combined with its parameter-free nature, makes \textbf{AE50} a more reliable choice for deployment across diverse environments.
\textbf{DRL} is the only method that consistently performs on par with or better than \textbf{AE50}.
As shown in \autoref{table:rpe}, the \ac{RTE} and \ac{RRE} distributions for \textbf{DRL} are significantly lower than those of \textbf{AE50}. 
The predictive nature of \textbf{DRL} enables optimized training for visual localization tasks, offering promising advancements for exposure control.
However, a key limitation of \textbf{DRL} is its lack of generalization due to its dependence on the training dataset, leading to domain shift when deployed in new environments~\citep{zhang2021adaptive}. 
Therefore, while \textbf{DRL} shows better accuracy in specific metrics, the robustness of \textbf{AE50} makes it a more versatile and reliable exposure control method overall.

\begin{table}
\caption{Statistical difference with \textbf{AE50} based on a \SI[detect-weight=true,mode=text]{95}{\percent} confidence Mann-Whitney U test. \textcolor{ForestGreen}{$\blacktriangle$}: Significantly better than \textbf{AE50}, \textcolor{Mahogany}{$\blacktriangledown$}: Significantly worse than \textbf{AE50}, and  $\equiv$ : Not significantly different from \textbf{AE50}.}
\setlength{\tabcolsep}{5pt}
\renewcommand{\arraystretch}{1.2}
\begin{tabular}{L{1.6cm}|cccccccc}
\hline
 Metric & \textbf{Fix} & \textbf{AE30} & \textbf{AE70} & \textbf{Shim} & \textbf{Wang} & \textbf{Kim} & \textbf{DRL} \\
\hline
Coverage & $\equiv$ & $\equiv$ & \textcolor{Mahogany}{$\blacktriangledown$} & $\equiv$ & \textcolor{Mahogany}{$\blacktriangledown$} & $\equiv$ & $\equiv$ \\
\# Matches & $\equiv$ & $\equiv$ & $\equiv$ & $\equiv$ & \textcolor{Mahogany}{$\blacktriangledown$} & $\equiv$ & $\equiv$ \\
\ac{RTE} & \textcolor{Mahogany}{$\blacktriangledown$} & \textcolor{Mahogany}{$\blacktriangledown$} & \textcolor{Mahogany}{$\blacktriangledown$} & \textcolor{Mahogany}{$\blacktriangledown$} & \textcolor{Mahogany}{$\blacktriangledown$} & \textcolor{Mahogany}{$\blacktriangledown$} & \textcolor{ForestGreen}{$\blacktriangle$} \\
\ac{RRE} & \textcolor{Mahogany}{$\blacktriangledown$} & \textcolor{Mahogany}{$\blacktriangledown$} & \textcolor{Mahogany}{$\blacktriangledown$} & \textcolor{Mahogany}{$\blacktriangledown$} & \textcolor{Mahogany}{$\blacktriangledown$} & $\equiv$ & \textcolor{ForestGreen}{$\blacktriangle$} \\
Failure Time & $\equiv$ & $\equiv$ & $\equiv$ & $\equiv$ & \textcolor{Mahogany}{$\blacktriangledown$} & $\equiv$ & $\equiv$ \\
\hline
\end{tabular}
\label{table:p_value_flipped}
\end{table}

\subsection{QUALITATIVE COMPARISON}

\begin{figure*}[htbp]
	\centering
	\includegraphics[width=0.98\textwidth]{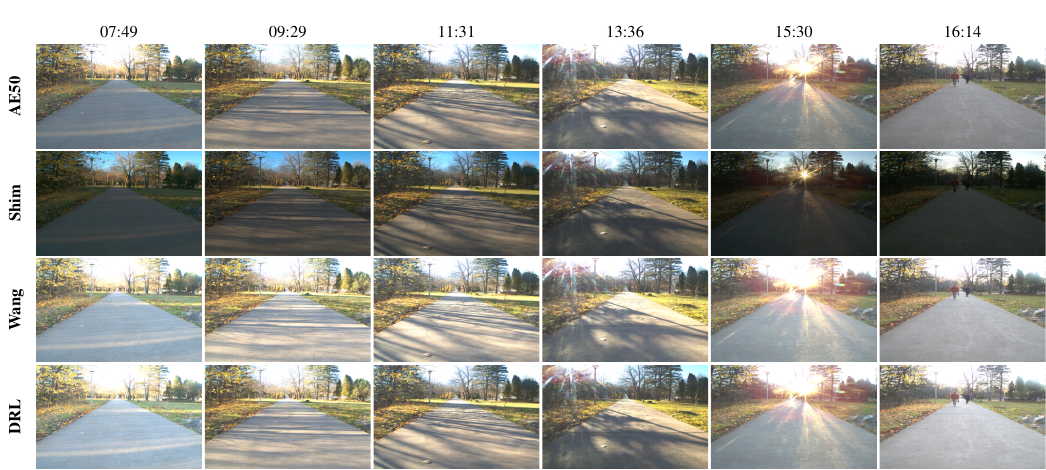}
	\caption{Qualitative comparison of the same location at different time of the day from the \textbf{Campus2} sequences.
    Each row is the emulated image for the method written on the left.
    Note the quality of the emulated images, which exhibit photorealistic characteristics.}
	\label{fig:results_exposure_comparison}
\end{figure*}

To conclude our evaluation of \ac{AE} methods, we present emulated images from our extended \datasetName~dataset in \autoref{fig:results_exposure_comparison}. 
These images are taken from the same location along the \textbf{Campus2} trajectory, with columns representing different times of the day and rows showing the emulated outputs from the top-performing methods.
Throughout the day, the illumination drastically alters the appearance of the outdoor scene, transitioning from no sunlight to direct sun glare, with shadows changing the appearance of the scene. 
The emulated images of \textbf{AE50} and \textbf{DRL}, the highest-performing methods, show very similar exposure levels, though \textbf{DRL} tends to favor slightly higher exposure times. 
Notably, \textbf{Shim} results in darker images compared to the others, which helps mitigate blur by avoiding overexposure, making it the most reliable method on the \textbf{Campus2} sequences, as shown in \autoref{table:success_trajs}.
Its use of lower exposure times helps it handle sun glare, as seen in the \textsc{15:30} sequence.

\section{LIMITATIONS AND LESSONS LEARNED}
\label{sec:lessons_learned}

In this section, we acknowledge the limitations of both our emulator and our backpack, drawing on the insights gained from field deployments and the development process. 
Additionally, we share lessons learned from building a custom acquisition platform from the ground up, with the aim of helping the community by addressing the challenges we encountered during this process.

\subsection{EMULATOR LIMITATIONS}
\label{sec:limitations_emulator}

Based on the thorough analysis of our emulator and the experiments conducted in ~\autoref{sec:results}, this section outlines the limitations of our emulation method. 
As detailed in \autoref{alg:pseudo_code}, our emulator operates based on a discrete decision scheme, meaning the selection criterion for $I_\text{source}$ is not a continuous function across the acquired brackets. 
Consequently, when $\Delta t_\text{target}$ is selected immediately before or after one of the collected image brackets, the $I_\text{source}$ used will differ between the two exposure times, resulting in images with noticeable differences even for closely spaced exposure times. 
While this behavior was observed qualitatively and made the sequence flow less aesthetically pleasing, it did not impact the performance of ORB-SLAM2.
A limitation of our emulator is its inability to manage camera gain. 
This constraint led the \ac{AE} methods to select high exposure times in an attempt to compensate for the absence of gain adjustment, whereas a typical camera system would balance gain and exposure time.
Additionally, the emulator does not account for motion blur, with the only motion blur modeled being the inherent blur from the acquired image brackets. 

\subsection{BACKPACK LIMITATIONS}
\label{sec:limitations_backpack}

A backpack offers simplified operation, but may lack some features commonly found in standard vehicles.
For instance, no odometry data is available because of the absence of wheel encoders, which can be useful for numerous autonomous navigation algorithms.
To compensate, the \ac{IMU} is used to estimate the velocity and the position by integrating the accelerometer values.
Since these values drift quickly, the \ac{IMU}'s odometry estimations are corrected using a 3D map built from the lidar point clouds.
The code is available on GitHub.\footnote{\url{https://github.com/norlab-ulaval/imu_odom}}

Maintaining a consistent speed during data collection is challenging due to human factors. 
It is particularly difficult to sustain a constant walking speed over long distances, especially when the desired pace is low. 
Fatigue, which is influenced by the weight of the platform, also affects the walking velocity. 
The platform's significant weight primarily results from the enclosure and aluminum structure, which are essential for ensuring resistance to harsh weather conditions and vibrations, thereby enabling more stable calibrations. 
However, the platform's weight can make it challenging or even impossible to wear, depending on the operator. 
Additionally, oscillations in the sensor trajectories are often observed due to walking movements, which can be problematic for certain experiments. 
To mitigate these limitations, one possible solution is to use the backpack as an acquisition head, mounted on a robotized platform. 
As shown in \autoref{fig:mosaic} panel G, we employed such a setup to collect the extension of \datasetName.

\subsection{LESSONS LEARNED}
\label{sec:lessons_learned_backpack}

\begin{figure*}[htbp]
	\centering
	\includegraphics[width=0.98\textwidth]{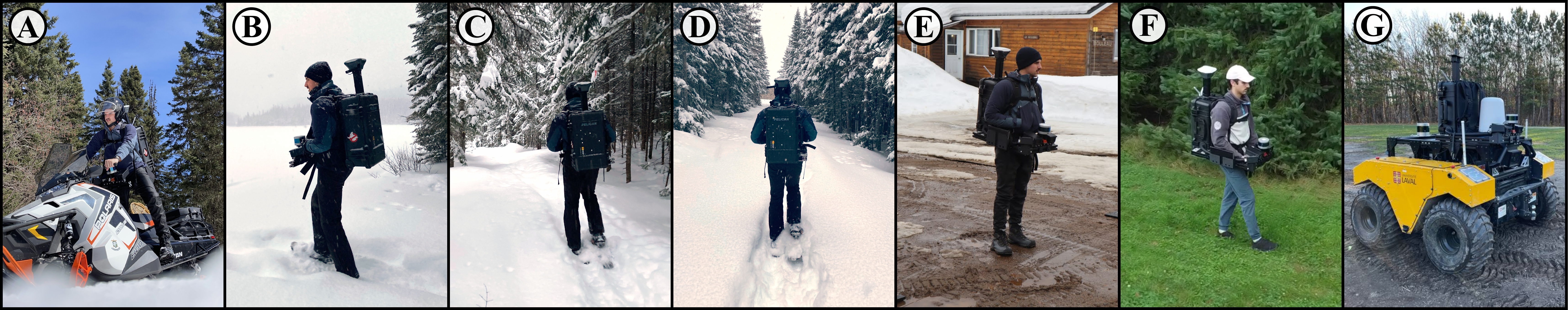}
	\caption{Examples of environments traveled with the backpack. \textbf{(A)} Winter displacement on a snowmobile, \textbf{(B)} Winter frozen lake, \textbf{(C)} Winter dense forest, \textbf{(D)} Winter tree corridor, \textbf{(E)} Spring muddy forest, \textbf{(F)} Summer forest, and \textbf{(G)} Backpack fixed on a Clearpath Warthog, which was used to record the \datasetName~extension.}
	\label{fig:mosaic}
\end{figure*}

This section provides insights into the development of a robotic acquisition platform, informed by the challenges encountered during the design of the backpack. 
These lessons were learned after the backpack was used over more than \SI{25}{\kilo\meter} across various environmental conditions, as shown in \autoref{fig:mosaic}. 
The backpack was deployed year-round, enduring harsh weather conditions including snowstorms, rain, and hot summer days. 
Additionally, it was utilized in three industrial demonstrations, further demonstrating its robustness and reliability.

\textbf{Bandwidth} - The first purpose of the backpack was a platform to record camera images easily.
Since the two selected industrial cameras communicate at a rate of \SI{1}{\giga b \per \second}, it was challenging to obtain the maximum \ac{FPS} given in the specification sheet.
The bandwidth was first limited by the switch, which had \SI{1}{\giga b \per \second} per port. 
The flux from the two cameras being \SI{2}{\giga b \per \second}, the transfer bandwidth from the switch to the Jetson Xavier AGX was not sufficient.
The effect was worse at first since the lidar was also plugged into the same switch, then it was moved to its independent Ethernet card.
After multiple attempts to optimize the communication channels in the switch itself, we came to the conclusion that the best way to develop a robust communication link from the cameras to the main computer was to increase the hardware performance.
By adding a \SI{10}{\giga b \per \second} Ethernet link, the cameras' \ac{FPS} is maximized without any packet loss.

\textbf{Plug-and-play System} - With the intention of not having to transport a computer during data gathering, we added a small LED screen and two buttons as described in \autoref{sec:experimental_setup_backpack_panel}.
The screen allows for real-time updates about the sensors and the computer, giving the user certainty about the recording information.
In our case, it is mostly useful to detect malfunctions, such as disconnected sensors, cameras overheating, lidar total obstruction, and \ac{GNSS}-denied environment.
Moreover, displaying the remaining free disk space allows for better time management, especially during long deployments.
An Android tablet provides real-time visualization of the images and point clouds, through the ROS-Mobile app,\footnote{\url{https://github.com/ROS-Mobile/ROS-Mobile-Android}} which would not be possible without the high-bandwidth Wi-Fi antenna.
This setup proved beneficial to adjust camera parameters in changing environment lighting, since it allows to validate the image quality before a new recording.
From our experiences, investing time upfront in a user-friendly platform allows for faster data gathering later on, while a status display provides a more robust and efficient acquisition.

\textbf{Camera Power Supply} - For outdoor data collection, industrial cameras should be protected from rain, dust, and snow using a housing.
However, by encasing them, the cameras will often overheat due to a lack of airflow.
This effect is enlarged when using \ac{PoE} as a power supply source for the cameras, since the provided voltage is higher than the cameras' rating. 
The excess is then converted to heat, which increases the ambient temperature surrounding the camera until it hits the protection threshold, causing lag, and even shutdown.
We found that powering the camera directly from a standard power cable, instead of using the \ac{PoE}, alleviates overheating problems.

\textbf{Camera Calibration Pipeline} - Camera intrinsic and extrinsic calibration is a complex process. 
To ensure accuracy, it is essential to collect a calibration dataset at the start of each recording session, as camera components and positioning are influenced by vibrations and temperature fluctuations. 
Since our cameras are optimized for outdoor performance, conducting the calibration indoors using the built-in \ac{AE} method yielded suboptimal results due to increased gain and/or exposure times.
To address this issue, we perform calibration with zero gain and a low exposure time. 
While these settings produce dark, visually unappealing images, they effectively eliminate noise from gain and prevent motion blur, even if the calibration board is slightly moving. This approach results in sharp images that the checkerboard detector can process reliably.
After capturing the dataset, we refine the calibration by manually selecting defect-free images. 
Using a \ac{RANSAC}-based algorithm, we compute the intrinsic and extrinsic parameters from a random subset of images, minimizing the reprojection error for improved accuracy.




\section{CONCLUSION}
Building on our previous work~\citep{gamache2024exposing}, we present a reproducible pipeline for emulating realistic images, leveraging our emulator and the extended multi-exposure dataset \datasetName.
In this study, we expanded \datasetName~by capturing six repetitions of the same trajectory throughout a full day, covering a wide range of illumination conditions.
The complete \datasetName~dataset now includes \numberTrajectories~trajectories for \kilometerTravelled\si{\kilo\meter} and totaling \totalHours~\si{\hour} of recorded data.
Using this dataset and emulator, we generated images for eight \ac{AE} methods and evaluated their performance across several metrics relevant to visual \ac{SLAM}.
Our findings make us concludes that the classical method, \textbf{AE50}, remains a robust and versatile choice due to its simplicity.
However, we highlight predictive approaches, such as \textbf{DRL}, as particularly promising for future development, given their capability to estimate optimal exposure settings for subsequent frames.
Our emulator-based approach facilitates fully reproducible offline testing and allows additional \ac{AE} methods to be incorporated into the benchmarks without the need for new data collection.
To further support reproducibility, we provide comprehensive documentation of our equipment setup, including insights gained from deploying our backpack over more than \SI{25}{\kilo\meter} across all seasons.
In the future, enhancing the emulator to emulate additional factors such as sensor gain and motion blur could increase its realism and utility for broader applications.





\printbibliography

\begin{IEEEbiography}[{\includegraphics[width=1in,height=1.25in,clip,keepaspectratio]{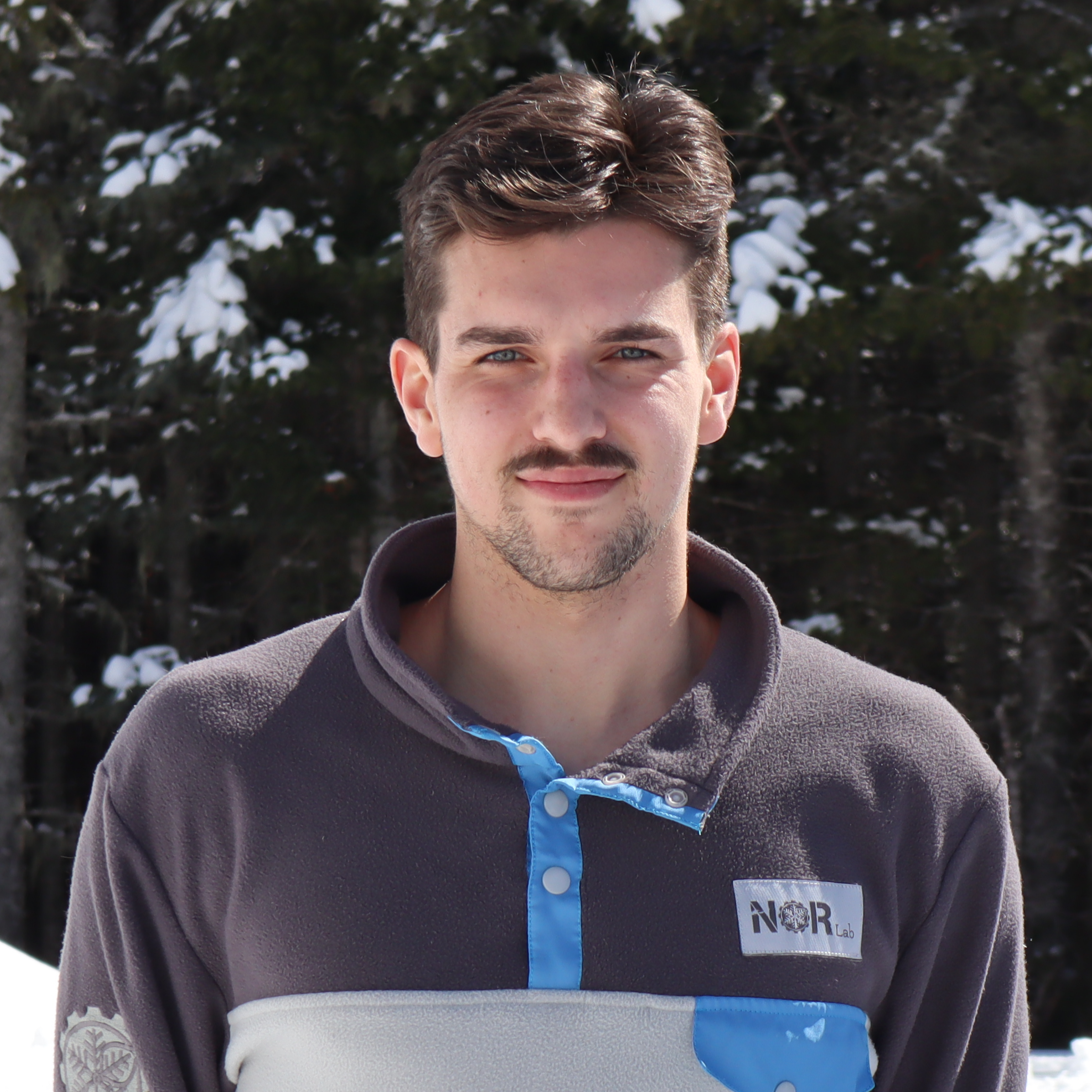}}]
{OLIVIER GAMACHE}~received the B.Eng. degree in physics at Polytechnique Montréal (Québec, Canada) in 2020. 

He is currently Ph.D. student in computer science after an accelerated passage from the MS in mobile robotics at Université Laval (Québec, Canada). His research interests are related to computer vision for robotics, visual localization and mapping, and field deployment of robotic systems. 
\end{IEEEbiography}

\begin{IEEEbiographynophoto}
{JEAN-MICHEL FORTIN,}~photograph and biography not available at the time
of publication.
\end{IEEEbiographynophoto}

\begin{IEEEbiographynophoto}
{MAT\v EJ BOXAN,}~photograph and biography not available at the time
of publication.
\end{IEEEbiographynophoto}

\begin{IEEEbiography}
[{\includegraphics[width=1in,height=1.25in,clip,keepaspectratio]{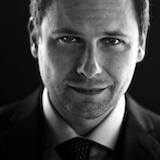}}]
{FRANÇOIS POMERLEAU}~(Senior Member IEEE), received the Ph.D. degree in mechanical engineering from the Autonomous Systems Lab at ETH Zurich (Switzerland) in 2013, and the M.Sc. degree in electrical engineering from IntroLab at Université de Sherbrooke in 2009. 

He is professor at Université Laval since 2017. His research interests include 3D reconstruction of environments using laser data, autonomous navigation, search and rescue activities, environmental monitoring, trajectory planning and scientific methodology applied to robotics. 

Prof. Pomerleau is professional Member of the Order of Engineering of Quebec (OIQ), Distal fellow of the NSERC Canadian Robotics Network (NCRN), Member of the NSERC-CREATE program on collaborative robotics (CoRoM), Member of Research center of robotics, vision and artificial intelligence (CeRVIM), and Member of Strategic center of distributed intelligent and shared environments (REPARTI). 
\end{IEEEbiography}

\begin{IEEEbiographynophoto}
{PHILIPPE GIGUÈRE,}~photograph and biography not available at the time
of publication.
\end{IEEEbiographynophoto}

\vfill\pagebreak

\end{document}

%% file: latexGoodPractices/preamble.tex
\usepackage[l2tabu,orthodox]{nag}

\usepackage[
    backend=bibtex8,
    style=ieee,
    sorting=none,
    natbib=true,
    doi=false,
    isbn=false,
    url=false,
    eprint=false,
    maxcitenames=1,
    mincitenames=1
]{biblatex}

\usepackage[pdftex,colorlinks]{hyperref}

\usepackage[french,english,noabbrev,nameinlink]{cleveref}

\usepackage[printonlyused]{acronym}

\usepackage{siunitx}
\usepackage[all]{nowidow}

\usepackage[dvipsnames]{xcolor}

\usepackage{lipsum}

\usepackage{xspace} 
\newcommand{\ie}{i.e.,\xspace{}}

\usepackage[pdftex]{graphicx}

\usepackage{epstopdf}

\usepackage{import}

\graphicspath{{./latexGoodPractices/}}

\usepackage{booktabs}

\usepackage{tabularx}
\usepackage{multirow, multicol}

\usepackage{amssymb,amsfonts,amsmath,amscd}

\usepackage{bm}

\newcommand{\bbm}{\begin{bmatrix}}
\newcommand{\ebm}{\end{bmatrix}}